\documentclass[final,5p,times,twocolumn]{elsarticle}
\usepackage[utf8]{inputenc}

\usepackage{amssymb}
\usepackage{amsmath,bm}
\usepackage{algorithm}
\usepackage{algpseudocode}
\usepackage{caption}
\usepackage{graphicx} 
\usepackage{url}
\usepackage{array}
\usepackage{makecell}
\usepackage{booktabs}
\usepackage[table]{xcolor}
\newcolumntype{P}[1]{>{\centering\arraybackslash}p{#1}}
\usepackage{subcaption}
\usepackage{multirow}
\usepackage{comment}
\usepackage{soul}

\journal{.}

\usepackage{amsfonts}
\usepackage{amsthm}
\usepackage{amsfonts}
\usepackage{amsmath}
\usepackage{amscd}
\usepackage{bbm}
\usepackage{booktabs}

\usepackage{t1enc}
\usepackage[mathscr]{eucal}
\usepackage{indentfirst}
\usepackage{graphicx}
\usepackage{xcolor}
\usepackage{hyperref}
\usepackage{graphics}
\numberwithin{equation}{section}
\usepackage{hyperref}
\usepackage{algorithm}
\usepackage{algpseudocode} 
\usepackage{footnote}
\makesavenoteenv{figure}

\begin{document}

\begin{frontmatter}
 
\title{Evaluation of LLMs for mathematical problem solving}

 \author[2]{Ruonan Wang \corref{contrib}}
\author[2]{Runxi Wang \corref{contrib}}
\author[2]{Yunwen Shen\corref{contrib}}
\author[2]{Chengfeng Wu \corref{contrib}}
\author[2]{Qinglin Zhou\corref{contrib}}
\author[2]{Rohitash Chandra \corref{corauthor}}


 
\affiliation[2]{Transitional Artificial Intelligence Research Group, School of Mathematics and Statistics, UNSW Sydney, Sydney, Australia}

\cortext[corauthor]{Corresponding author}
\cortext[contrib]{Authors contributed equally. The order of the author names is randomly listed.}

\begin{abstract}
Large Language Models (LLMs) have shown impressive performance on a range of educational tasks, but are still understudied for their potential to solve mathematical problems. In this study, we compare three prominent LLMs, including GPT-4o, DeepSeek-V3, and Gemini-2.0, on three mathematics datasets of varying complexities (GSM8K, MATH500, and University datasets). We take a five-dimensional approach based on the Structured Chain-of-Thought (SCoT) framework to assess final answer correctness, step completeness, step validity, intermediate calculation accuracy, and problem comprehension. The results show that GPT-4o is the most stable and consistent in performance across all the datasets, but particularly it performs outstandingly in high-level questions of the University dataset. DeepSeek-V3 is competitively strong in well-structured domains such as optimisation, but suffers from fluctuations in accuracy in statistical inference tasks. Gemini-2.0 shows strong linguistic understanding and clarity in well-structured problems but performs poorly in multi-step reasoning and symbolic logic. Our error analysis reveals particular deficits in each model: GPT-4o is at times lacking in sufficient explanation or precision; DeepSeek-V3 leaves out intermediate steps; and Gemini-2.0 is less flexible in mathematical reasoning in higher dimensions. 

\end{abstract}



\end{frontmatter}

%
\section{Introduction}

Solving mathematical problems \cite{liljedahl2019mathematical} is one of the major challenge for students \cite{tambychik2010students} in the current educational system. This difficulty mainly stems from the abstraction \cite{mitchelmore2004abstraction} and strict logical frameworks \cite{kinard2008rigorous} that are intrinsic to mathematics, and the extremely wide field of its application areas. Students need to have a strong theoretical background and be able to apply mathematical concepts to situations in daily life. However, most students are lacking in various areas of learning mathematics, especially in conceptual understanding\cite{yushau2004mathematics}, computational anxiety\cite{hembree1990nature} and the mastery of  applying mathematical concepts flexibly. Mathematics requires the acquisition of advanced abstract \cite{susac2014development} and logical thinking skills\cite{ab2019logical}, as well as the mastery of a large number of formulas and theorems. However, the current education system aims at providing theoretical knowledge without sufficiently strengthening  the practical skills and critical thinking\cite{firdaus2015developing} ability of student. Students tend to passive recipients of information with limited opportunities for active exploration and critical analysis\cite{peter2012critical}. It is difficult for students to apply their knowledge to practical problems effectively. Therefore, enriching the traditional teaching process to build up students' practical work ability and thinking capacity is an significant issue in mathematics education.

In addition, advances in computer and network technology \cite{bin2022artificial} have driven the application of educational technology (EdTech)\cite{lazar2015importance} in mathematics teaching. The emergence of educational technology has transformed traditional teaching models \cite{voskoglou2020benefits}. By using varied pedagogies and active learning experiences, educational technology improves students' learning of mathematics and their ability to problem-solve \cite{telegina2017use}. Furthermore, online learning systems \cite{larreamendy2006going} expose students to a wide range of resources and interactive lessons through which they can enhance their understanding of concepts and improve independent learning strategies.

Progress in natural language processing (NLP) ~\cite{zubiaga2024natural} based on artificial intelligence technologies~\cite{gao2021introduction} and the advent of large language models (LLMs) ~\cite{matzakos2023learning,johnsen2024large,shanahan2024talking} have led to a revolutionary change in educational technology, giving rise to novel possibilities for mathematics education. The application of LLMs ~\cite{hadi2023survey,zhao2023survey,kaddour2023challenges} has fundamentally changed the design of human-computer interaction in educational settings ~\cite{vemuri2024evolution,kothari2024enhancing} and initiated a revolutionary change in instructional approaches. In contrast to conventional systems that rely on predefined rules, large language models (LLMs) demonstrate more advanced language comprehension and generation capabilities ~\cite{hadi2023large}. Not only do they accurately comprehend user intent and produce natural, fluent text content ~\cite{johnsen2024large}, but they also adapt their output dynamically to feedback, thus creating a more semantically rich and interactive learning experience ~\cite{laleh2024survey}. These characteristics create new opportunities for applying LLMs and it has significantly driven the evolution of human-machine collaboration models in multiple fields ~\cite{puerta2025multifaceted}, including educational tutoring, customer service, programming assistance, and interactive communication support, enabling them to achieve better knowledge acquisition and problem-solving in learning, work.
Large language models have demonstrated significant potential in mathematics education. Students have widely utilised LLMs to obtain problem-solving \cite{matzakos2023learning} strategies and answers \cite{matzakos2023learning},  significantly shortening the time required to brainstorm and solve problems. However, whether this has truly improved learning outcomes remains to be explored \cite{yan2024practical}. Additionally, LLMs have the ability to accurately analyse the student's learning situations\cite{luan2020challenges} and dynamically adjust the presentation of content based on their pace and preferences\cite{luan2020challenges}, which can help promote "anytime, anywhere learning" teaching methods\cite{karlstrom2024exploring}. As LLM technology continues to advance, the potential for its use in education is becoming more widespread. The deep integration of mathematics education with artificial intelligence \cite{opesemowo2024artificial} is opening up new learning pathways and driving the construction of efficient, personalised learning platforms, thereby improving teaching quality and efficiency and promoting the comprehensive development of students\cite{tan2022information}.
Modern LLMs are typically built on deep learning models \cite{lecun2015deep,wu2018development}  and widely applied in NLP tasks \cite{kumar2024large}\cite{chen2024large}. These models demonstrate strong capabilities in text understanding and generation capabilities and are widely applied in machine translation, text summarisation, and code generation \cite{jiang2024survey}. 
Their task versatility stems from large-scale pretraining \cite {liu2024datasets} and their capacity to manage tasks in both zero-shot and few-shot learning \cite{meshkin2024harnessing}. LLMs include GPT-4o \cite{alto2023modern},   Gemini-2.0\cite{imran2024google}, and DeepSeek-V3\cite{neha2025survey} capable of text generation, logic reasoning, and multilingual processing\cite{rahman2025comparative}. Despite their advantages, LLMs encounter several challenges, such as hallucination (i.e., generating false information\cite{hadi2023survey}, and inherent data biases reflecting social and cultural stereotypes \cite{gallegos2024bias}\cite{thapa2025large}\cite{myers2024foundation}\cite{kamath2024llm}. 
Although current LLMs demonstrate outstanding capabilities in mathematical problem solving, formula derivation, and theorem explanation \cite{hosseinpour2018step}, they remain weak in symbolic computation, multi-step algorithms and formal theorem proving \cite{lai5002356solving}. To address these shortcomings, researchers have developed tools such as Wolfram Alpha\cite{davis2023testing} to enhance computational accuracy and logic rigour, and have improved programming performance through models like GitHub Copilot\cite{li2024assessing}, GPT, and Code Llama\cite{bhattacharya2023exploring}. However, these models still face challenges such as a lack of logical consistency\cite{lyu2024automatic}, difficulty handling large-scale dependency structures, and generating errors or fictional functions when dealing with complex tasks. Algorithmic breakthroughs are still required to meet the demands of complex application scenarios \cite{zhu2023intelligent}, and how to efficiently utilise LLMs to improve learning outcomes while ensuring fairness \cite{dai2019bridging} has become a focal point of current academic research.
Therefore, it is necessary to conduct systematic evaluations of LLMs using mathematical benchmark \cite{liu2024mathbench} datasets such as GSM8k \cite{didolkar2024metacognitive} and MATH, which are widely used in mathematical tasks. And in order to study the performance of LLMs in advanced-level mathematical problems, we plan to construct the University of New South Sales problem dataset, which is collected by using assignment and final exams from the Mathematics and Statistics School. These evaluations help us to identify the strengths and weakness of models across different tasks \cite{laskar2023systematic},  and provide empirical support for model optimisation. Although existing studies have fully demonstrated the ability of LLMs in mathematical and programming tasks, their shortcomings in logical reasoning and computational accuracy\cite{chang2024survey} still limit their widespread application.  Therefore, it is imperative to continue optimising LLMs from the perspectives of algorithm design and technical pathways to enhance their overall performance and meet the increasingly complex and diverse demands of practical applications\cite{surampudi2024big}.
Our study is based on this premise and aims to evaluate the performance of three mainstream LLMs, GPT-4o, DeepSeek-V3, and Gemini-2.0, in solving mathematical problems. We will examine their capabilities in terms of accuracy, reasoning quality, completeness, and consistency using a set of problems covering different levels of difficulty. We will use an automated quantitative evaluation framework to provide theoretical support and practical evidence for their application in education.
Our study will focus on the following contributions:
\begin{itemize}
\item Systematically evaluate the accuracy and efficiency of diverse LLMs in solving mathematical problems with different levels of complexity.
\item Compare the strengths and weaknesses of each model in terms of mathematical reasoning, accuracy and correctness of solutions, and problem-solving adaptability.
\item Provide advice about different types of LLMs for mathematical problems that can be used to develop innovative approaches in the development of assignments and exams in mathematics education in secondary schools and universities.
\end{itemize}

\section{Related Work and Background}

As the application of large language models in mathematics education continues to expand, questions regarding their capabilities, impacts, and suitability have increasingly drawn academia attention. As previously mentioned that LLMs have not only transformed the way humans interact with machines but also provided new tools and approaches for solving mathematical problems. However, their performance in actual educational settings has also been met with controversy and challenges. This section aims to review relevant research, starting from the issue of academic integrity, to conduct an in-depth analysis of the technological evolution of LLMs and their performance in mathematical tasks, thereby providing theoretical support for future empirical evaluations.

\subsection{Academic Integrity and the Boundaries of AI Use}

Although LLMs demonstrate outstanding performance in terms of generative capabilities and problem-solving efficiency, their tendency to "directly provide answers" has raised concerns about academic integrity\cite{castillo2023effect}. In an academic environment, students may bypass the thinking process and rely directly on models to complete assignments or exams. This phenomenon not only undermines students' self-directed learning but also may lead to academic misconduct\cite{zhai2024effects}. In addition, due to the easy accessibility of LLMs, the risk of their misuse for online cheating in examination scenarios increases significantly, which seriously threatens the validity of traditional assessment methods\cite{susnjak2024chatgpt}. Therefore, how to balance the use of technology with learning outcomes has become a focal point of current discussions.\cite{nguyen2023ethical}. At the same time, in recent years, some researches on the capabilities of LLMs in mathematical and programming capabilities is promoted with the initial aim of improving the quality of computer-assisted learning\cite{yang2024formal}. As an educational tool, LLMs can personalise content, provide real-time feedback, and accelerate the learning process. In the field of scientific research, LLMs can reduce the burden of researchers in routine tasks, so that they can devote more energy to high-level innovative work\cite{alto2023modern}. The future development direction should focus on improving the explanatory ability and logical reasoning level of the model, as well as improving the evaluation system and anti-cheating mechanism to ensure the suitable and responsible application of artificial intelligence in education\cite{monib2024generative}\cite{motlagh2023impact}.

\subsection{Model Selection and Technical Foundations}

The development of large language models is based on breakthroughs in deep learning \cite{janiesch2021machine} technology. In particular, neural network ~\cite{rusk2016deep,wang2024scaling} models with the Transformer architecture at their core have enabled language processing systems to model complex contextual relationships ~\cite{thota43deep,hao2016deep} more efficiently. Compared to traditional machine learning \cite{mehrotra2019basics} methods that rely on manually designed features  \cite{singh2023study}, such as linear regression and support vector machines \cite{liu2011supervised}, deep learning uses multi-layer network structures to automatically extract multi-level semantic representations from large-scale data, significantly improving the generalisation and task adaptability of models.

In the field of natural language processing (NLP), the introduction of the Transformer architecture \cite{luo2023self} marked the entry of generative language models into practical use \cite{raiaan2024review}. Through large-scale pre-training language modelling tasks, models can learn language structures, semantic relationships, and even a certain degree of reasoning ability. Subsequently, combined with supervised fine-tuning \cite{juhasz2024large} and human feedback optimisation mechanisms (such as RLHF), large language models began to acquire the ability to execute complex instructions and generate multi-step reasoning processes\cite{rane2024machine}. In recent years, various models such as the GPT series, DeepSeek system, and Gemini framework have been released based on this technical framework and have achieved significant results in tasks such as language understanding, code generation, and mathematical modelling \cite{lecun2015deep}.

Our study selected three representative models-GPT-4o, DeepSeek-V3, and Gemini-2.0-as the subjects for evaluating mathematical problem-solving capabilities. The following sections will introduce their technical features and task performance, providing a foundation for subsequent method and result analyses.

\subsubsection{ChatGPT and GPT-4o}

GPT developed by Open AI, which is an LLM that can solve mathematical problems\cite{roumeliotis2023chatgpt,raiaan2024review} based on the Transformer model\cite{yenduri2024gpt}. It mainly introduces the self-attention mechanism\cite{wang2022research} to achieve a global modelling of word relationships in the input text and then profoundly understands and generates the context. Therefore, the model can understand the structure of mathematical expressions, which helps solve complex mathematical problems.  GPT-based LLMs have been pre-trained using a self-supervised learning objective \cite{wang2024chatgpt} on large-scale text corpora, enabling them to acquire general language understanding and preliminary reasoning capabilities.  Reinforcement Learning from Human Feedback (RLHF) \cite{kaufmann2023survey} has been used to fine-tune the model through instruction-based learning, enabling it to generate outputs that better align with human preferences in language style and logical expectations when solving mathematical problems. GPT has performed well in NLP \cite{liu2023summary} and can efficiently complete various complex tasks, such as dialogue generation, text writing, logical reasoning, and programming assistance \cite{liu2023external}. It provides efficient help at the exploratory, heuristic, and teaching levels, and it is a powerful auxiliary tool for research, teaching, and mathematics learning.

Multiple studies have indicated that GPT demonstrates a certain degree of generalisability in mathematical tasks, particularly in problems with clear structures~\cite{daher2024use,dao2023investigating,sanchez2023chatgpt,frieder2023mathematical}. It can handle basic computational and symbolic reasoning problems, such as algebraic operations, equation solving, function analysis, and limit calculations, and typically provides clear step-by-step derivations. In modelling and application-oriented tasks, GPT also demonstrates some auxiliary capabilities, assisting in the construction of mathematical models for statistical, probabilistic, and optimisation problems, and attempting to explain the underlying principles involved. In programming and numerical computation, the model can generate basic code in languages such as Python, Matlab, Stata, and R based on task requirements, supporting numerical simulation and data analysis. Notably, GPT demonstrates strong understanding capabilities when converting natural language problems into mathematical expressions, providing structured problem-solving pathways. Additionally, it is suitable foe explaining complex mathematical concepts such as linear algebra, differential equations, and number theory, thereby enhancing learners' understanding of mathematical principles.

\subsubsection{DeepSeek}

DeepSeek is a general-purpose large language model (LLM) independently developed in China \cite{youvan2025deepseek} , designed to support bilingual understanding, logical reasoning, code generation, and mathematical modelling. Since its initial release in 2023, the DeepSeek team has iteratively advanced the model architecture, successively launching DeepSeek-V1, V2,and V3 ~\cite{liao2025deepseek}, in an effore to establish a scalable, controllable, and continually evolving Chinese LLM framework \cite{puspitasarideepseek}.

The earliest version, DeepSeek-V1 \cite{puspitasarideepseek} followed the standard Transformer architecture and demonstrated excellent natural language instruction-following capabilities \cite{puspitasarideepseek}. It could handle basic arithmetic operations, text problem analysis, and algebraic equation solving tasks. However, its performance in multi-step reasoning tasks was limited, with frequent inconsistencies and a lack of support for symbolic computation or tool-based workflows\cite{he2025survey,neha2025survey}.
 
DeepSeek-V2\cite{puspitasarideepseek} introduced the Mixture of Experts (MoE) architecture \cite{gan2025mixture,tie2025survey}, which improves inference efficiency while maintaining parameter scale, and enhanced its ability to handle math and programming tasks through the sub-model DeepSeek-MATH \cite{miao2024ai}. This version begins to support Chain-of Thought (step-by-step reasoning) \cite{wang2025dynamic} output and can invoke Python/SymPy \cite{shao2024deepseekmath} and other tools to complete symbolic computation. It demonstrates good performance in solving mid-to-high-level mathematical problems on the GSM8K, MATH, and SVAMP \cite{park2024ensembling} datasets, particularly in tasks involving functions, geometry, and algebra.

DeepSeek-V3\cite{puspitasarideepseek} returns to a fully parameterised dense architecture to further improve language logical consistency, cross-task generalisation, and accuracy in Chinese and English expression \cite{liu2024deepseek}. This version shows significantly enhanced stability in multi-round question-answering, scientific writing, and complex mathematical problem handling, with improved text understanding and ambiguous problem resolution capabilities \cite {wang2025review}. Additionally, DeepSeek-V3 demonstrates strong explanatory reasoning capabilities, clearly presenting reasoning chains, and has achieved performance comparable to GPT-4-Turbo in code invocation and formula generation. It shows good consistency in a bilingual environment and can collaborate with subsystems such as DeepSeek-Coder \cite{guo2024deepseek} and DeepSeek-VL \cite{liao2025deepseek} to support mixed-media mathematical tasks (e.g., chart questions and geometric recognition).

Overall, the DeepSeek model demonstrates continuous evolution in mathematical language modelling, code generation, and symbolic reasoning. Its latest version has laid the foundation for participating in complex reasoning tasks, though there is room for improvement in flexibly addressing open-ended questions and adjusting strategies.

\subsubsection{Gemini}

Gemini is an LLM developed by Google DeepMind\cite{saeidnia2023welcome}, optimised for scientific computing and multi-model reasoning scenarios. It demonstrates strong capabilities in science and engineering,including mathematical problem solving  ~\cite{rane2024gemini}. Gemini integrates logical reasoning, symbolic computation, and code generation modules, supporting a wide range of applications from basic question types to complex scientific research problems. In comparison to other mainstream LLMs, Gemini places greater emphasis on the completeness of the reasoning chain and process interpretability \cite{xiang2025towards}. 

In terms of mathematical code generation, Gemini also performs stably, supporting mathematical libraries such as Python, NumPy, and SymPy \cite{campesato2024google}, and capable of completing tasks such as function plotting, symbolic solving, and numerical simulation. Its integrated "language-formula-program" \cite{zhang2025or} output method facilitates the complete closed loop from natural language questioning to mathematical modelling to program verification, making it particularly suitable for automated preliminary modelling of scientific research problems.


 


\section{Methodology}

We first present the framework used to evaluate selected LLMs for mathematical problem-solving. Our approach combines multiple benchmark datasets along with University assignments and exams. We incorporate controlled prompting strategies, with both quantitative and qualitative evaluation.

We need to conduct a series of controlled experiments using standardised mathematical problems to evaluate the mathematical problem-solving capabilities of LLMs. Therefore, we select three mainstream benchmark datasets- GSM8K, MATH, and the problem sets of the University of New South Wales \cite{ono2024evaluating} to cover different difficulty levels and problem types. In addition, we explored the impact of different prompting strategies, including zero-shot, few-shot, and Chain-of-Thought (CoT)\cite{sprague2024cot}, to examine each model's ability to perform multi-step reasoning.

\subsection{Mathematical problems}

To ensure representativeness and reliability of the evaluation results, this study selected three sets of mathematical problems from different levels and backgrounds: GSM8K, MATH 500, and the graduate course question bank from a selected University. These sets cover a wide range of problem types, from basic arithmetic and algebraic operations to advanced statistical reasoning and optimisation, and are characterised by diverse structures and distinct language styles.

\begin{itemize}
    \item GSM8K: This dataset was released by OpenAI, which contains 8,500 application-based problems at the elementary and middle school levels. The language is colloquial, and the solution paths typically range from multiple steps, making it suitable for testing a model's ability to process basic mathematical language and logic.
    
    \item MATH500: This dataset has 500 different difficulty and domain questions selected from the MATH benchmark, covering algebra, sequences, geometry, functions, and other fields. The question stems are close to academic teaching materials, with long chains of reasoning, suitable for analysing multi-step thinking abilities.
    
    \item University Graduate Question Bank:  This dataset consists of exam and assignment questions covering probability inference, matrix decomposition, optimisation theory, and financial computation, which are from the MIT Open Courseware. The language is rigorous, and the expression style is academic. Some questions have open structures, conceptual explanations, and non-standard expressions. The details about this dataset can be shown in Table \ref{tab:data_university}.

\begin{table*}[htbp]
\centering
\caption{Summary of mathematical problem dataset from MIT Open Courseware.}
\label{tab:data_university}
\renewcommand{\arraystretch}{1.5}
\begin{tabular}{cllcc}
\toprule
\textbf{Subject} & \textbf{Code} & \textbf{Title} & \textbf{Assessment Type} & \textbf{Year}\\
\midrule
\multirow{4}{*}{{Optimisation}}
&16.323 & Principles of Optimal Control
& Final Exam & 2007\\
&15.093 & Optimisation Methods & Final Exam & 2003/2006/2008\\
&6.253 & Convex Analysis and Optimisation & Midterm Exam & 2010/2012\\
&18085 & Computational Science and Engineering & Exam & 2008\\
\midrule
\multirow{2}{*}{{Financial Computational Math}}
&18.330 & Introduction to Numerical Analysis & Homework & 2012\\
&18.336 & Numerical Methods of Applied Mathematics & Problem Set & 2009\\
\midrule
\multirow{4}{*}{{Statistical Inference}}
&18.440 & Probability and Random Variables & Final Exam & 2014\\
&18.655 & Mathematical Statistics & Midterm Exam & 2016\\
&15.085J & Fundamentals of Probability & Final Exam & 2018\\
&6.431 & Probabilistic System Analysis & Final Exam & 2006\\
\bottomrule
\end{tabular}
\end{table*}
\end{itemize}

\begin{table}[H]
\centering
\caption{Research data sources}
\label{tab:datasources}
\small
\begin{tabular}{lllp{3.1cm}}
\toprule
\textbf{Dataset}&\textbf{Size}&\textbf{Level}&\textbf{Source}\\
\midrule
GSM8K & ~5000 & Primary & OpenAI public dataset\\
MATH 500 & ~500 & Intermediate & Hendrycks et al.(2021)\\
\textcolor{black}{University Problem} & ~\textcolor{black}{140} & \textcolor{black}{Advanced} & \textcolor{black}{MIT Open Courseware}\\
\bottomrule
\end{tabular}
\end{table}

All questions have undergone standardised pre-processing, including English standardisation, formatting, symbol conversion, and removal of non-structural redundancy. Additionally, we manually labelled each dataset with its knowledge domain (e.g., functions, linear algebra), problem-solving type (computational, reasoning, conceptual), and difficulty level (primary, intermediate, advanced) to support subsequent hierarchical analysis. We summarise the source, quantity, and difficulty level of each dataset category in Table \ref{tab:datasources}.


We balanced the types and difficulty levels of questions during sampling to avoid assessment bias due to an imbalance in question types. For example, in MATH500, we avoided selecting questions from several categories while ensuring that all disciplines (such as statistics, optimisation, and financial mathematics) were covered in the dataset.
In addition to supporting automatic comparison of model output results, we have manually annotated the standard answers and mathematical expression structures for each question. For some questions, we have established tolerance rules (e.g., accepting equivalent expressions and allowing a certain margin of error for floating-point numbers.

\subsection{Automated LLM Evaluation Framework}

Our evaluation focuses on the main performance dimensions: accuracy of final answers, correctness of intermediate computations, and symbolic processing capabilities. We combine quantitative metrics with qualitative error analysis to gain a comprehensive understanding of the problem, classify common failure patterns, and highlight the advantages and limitations of each model.

In particular, we expand the existing benchmark by introducing a set of original high-level problems from the University of New South Wales curriculum. These problems cover multiple topics and require varying degrees of abstract thinking, logical reasoning, and practical application skills, thereby providing a more rigorous test of mathematical ability.

Although previous studies have examined the mathematical performance of LLMs, such as mathematical reasoning, researchers have limited their evaluations to single models or narrow problem domains assessments \cite{lewkowycz2022solving}. In contrast, our study provides a broader cross-model comparison across multiple datasets and evaluation strategies. We aim to provide more practical insights into the current capabilities and limitations of LLMs in mathematics education by incorporating international benchmarks and university-level datasets with the following details.
   \begin{itemize}
       \item GSM8K \cite{cobbe2021training} (Focuses on basic mathematical reasoning);
       \item MATH 500\cite{hendrycks2021measuring} (Contains medium to high difficulty questions);
       \item \textcolor{black}{University Problem dataset (real academic assessment scenarios} 
    \end{itemize}

\begin{figure*}[htbp!]
\centering
\includegraphics[width=0.95\linewidth]{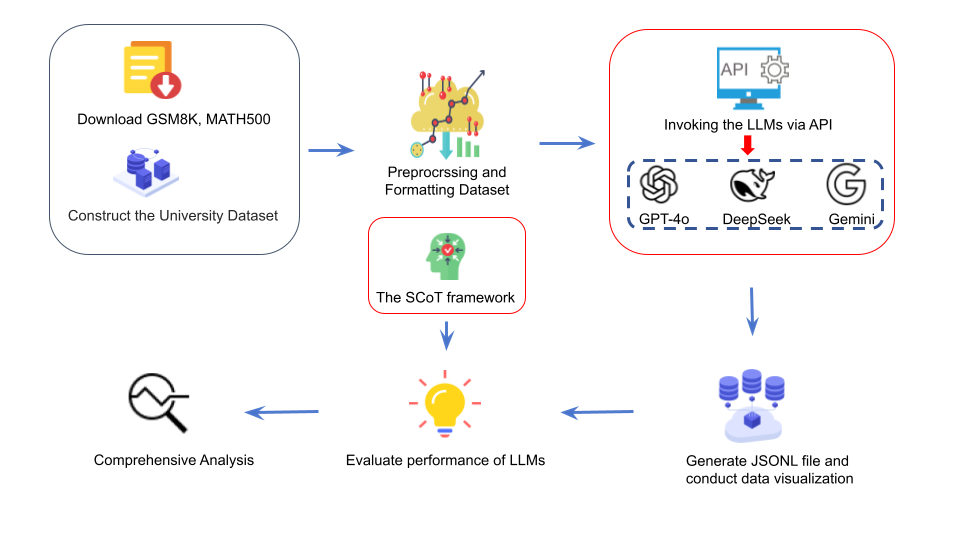}
\caption{Automated framework for evaluation of LLMs in mathematical problem-solving.}
\label{fig:LLMs_evaluation_workflow}
\end{figure*}

 We designed an automated evaluation framework (Figure \ref{fig:LLMs_evaluation_workflow}) to achieve reproducible evaluation of multiple LLMs on thousands of mathematical problems.  The overall design emphasises systematic and practical approaches, aiming to construct a fair, transparent, and reproducible evaluation framework\cite{chang2024survey} to get a better understanding of the actual performance of different  LLMs in solving mathematical problems \footnote{\url{https://docs.google.com/presentation/d/1CJbzXxncH7OPakhvoCSlXwrQ0s5JOyt7N3s_SitkQ40/edit?usp=sharing}}.


\begin{itemize}
    \item Stage 1 - Problem definition: We establish the foundations of the evaluation using the SCoT (Structure Chain-of-Thought (SCoT) \cite{lee2024applying} framework that emphasises systematic evaluation of the completeness and logical consistency of the reasoning chain to test the performance of LLMs, which serves as the basis for the five subsequent evaluation indicators design. 
    
    \item Stage 2 - Dataset pre-processing:
    We select many mathematical questions from the GSM8K, MATH500, and University datasets (Table \ref{tab:datasources}), covering multiple levels from basic arithmetic to advanced reasoning. After pre-processing, we classify the questions according to knowledge points and difficulty levels to facilitate subsequent layered evaluation and error analysis. We need to balance question types to ensure that different kinds of problems are represented in the evaluation.

    \item Stage 3 - LLM Invocation:  We invoke three language models- GPT-4o, DeepSeek-V3, and Gemini-2.0 via API interfaces to answer the given questions. We use a consistent prompt structure and prompt each question multiple times to verify the consistency of the model outputs. Furthermore, we verify the original model responses, timestamps, and other metadata as the basis for subsequent quantitative analysis and error classification,as shown in Table \ref{tab:evaluation_methods}.
   
    \item Stage 4 - Result Evaluation:  We review the model output results using error metrics from the literature  \cite{varastehnezhad2024llm} as shown in Table \ref{tab:evaluation_SCoT}.
    
\begin{table*}[htbp]
\centering
\caption{The Methods of the Evaluation for LLMs }
\label{tab:evaluation_methods}
\renewcommand{\arraystretch}{1}
\begin{tabular}{|p{3cm}|m{2.5cm}|p{12cm}|}
\toprule
\textbf{Category} & \textbf{Objection} & \textbf{Description}\\
\midrule
\multirow{2}{*}{Qualitative Analysis}
& Difficulty Level & The aim is to study the performance of large language models in solving mathematical problems of different difficulty levels within the same discipline.\\
& Domain(Subject) & The aim is to analyze the performance of large language models in solving mathematical problems under the same difficulty level across different disciplines.\\
\midrule
\multirow{3}{*}{Quantitative Analysis}
& GSM8K & A quantitative assessment of 5000 questions from GSM8K was conducted, with a focus on the correctness of the final answers and the rationality of the steps.\\
& MATH500 & A quantitative assessment was conducted on 500 questions selected from the MATH benchmark test to verify the performance of the language model in the multi-step reasoning process.\\
& University & \textcolor{black}{A quantitative analysis was conducted on 140 mathematical problems from various subjects from the MIT Open Courseware to study the performance of language models in advanced mathematics.}\\
\midrule
\multirow{2}{*}{Control Variables}
& Problem Type & Types include proof-based, conceptual, procedural, and word problems.\\
& Difficulty Level & All the questions were manually classified into 5 levels based on the length of the questionnaires. This classification controlled the variable of task difficulty, making the model more scientific when comparing performance horizontally.\\
\bottomrule
\end{tabular}
\end{table*}

\begin{table*}[htbp]
    \centering
    \caption{Evaluation dimension description based on the SCoT framework.}
    \renewcommand{\arraystretch}{1.4}
    \begin{tabular}{p{5.5cm}p{12cm}}
        \toprule
        \textbf{Dimension} & \textbf{Explanation} \\
        \midrule
        Accuracy & Check if the output answer is consistent with the correct answer, accepting floating point errors and equivalent expressions \\
        Reasoning Quality & Check if the explanation process is logically clear and the steps are reasonable \\
        Completeness & Check if all sub-questions can be answered based on both explicit and implicit conditions when generating results, and whether there are any skipped steps\\
        Consistency & Check if the model outputs consistent results for the same question across multiple runs \\
        Accuracy of Intermediate Calculations & Check if  each intermediate value is correct and free of calculation or clerical errors \\
        Understanding of the Question & Check if the model correctly interprets key conditions and understands what is being asked \\
        \bottomrule
    \end{tabular}
    \label{tab:evaluation_SCoT}
\end{table*}

    These evaluation dimensions refer mainly to the principles proposed by the SCoT framework and combine the types of problems we observed in our error analysis. Based on the automatic judgment of whether the model answers are correct, we manually analysed the model outputs for some representative questions to identify common failure patterns and logical defects \cite{kamoi2024evaluating}. We will brainstorm and reach a consensus on boundaries or uncertain answers to avoid being more subjective in grading. Qualitative analysis\cite{onwuegbuzie2014exemplar} was considered to observe the answering style, language expression, and development of ideas of different models. This can help us to gain an in-depth understanding and horizontal comparison of model behaviour.
    
\end{itemize}

\subsection{Evaluation Metrics and LLM Prompting Strategies}

The LLM evaluation focuses on two aspects: algebraic expression computation and mathematical reasoning problems \cite{gupta2025beyond}. Since LLMs are generative models, ensuring that the generated answers are accurate becomes very tricky, especially for mathematical reasoning tasks \cite{rane2023enhancing}. In the case of computational problems, we prefer to provide standard reference answers so that we can easily compare the model-generated answers to the correct answers. This facilitates the assessment of the model's performance in terms of numerical accuracy. We adopt the automatic proof verification method for the mathematical proof problem. Instead of relying on manual evaluations, we test the performance of the model by establishing the SCoT framework and an automatic scoring method inspired by recent advances in LLM-based reasoning verification \cite{ren2023self}. It provides a more objective criterion for evaluating mathematical reasoning problems and improves the efficiency of the evaluation process \cite{lu2024proof}. We also provide a prompting mechanism to assess the ability to correct their mistakes. If a model provided an incorrect answer, it would be prompted again to solve the problem, and the corrected answer would be evaluated.





\begin{figure*}[htbp]
    \centering
    \includegraphics[width=.81\linewidth]{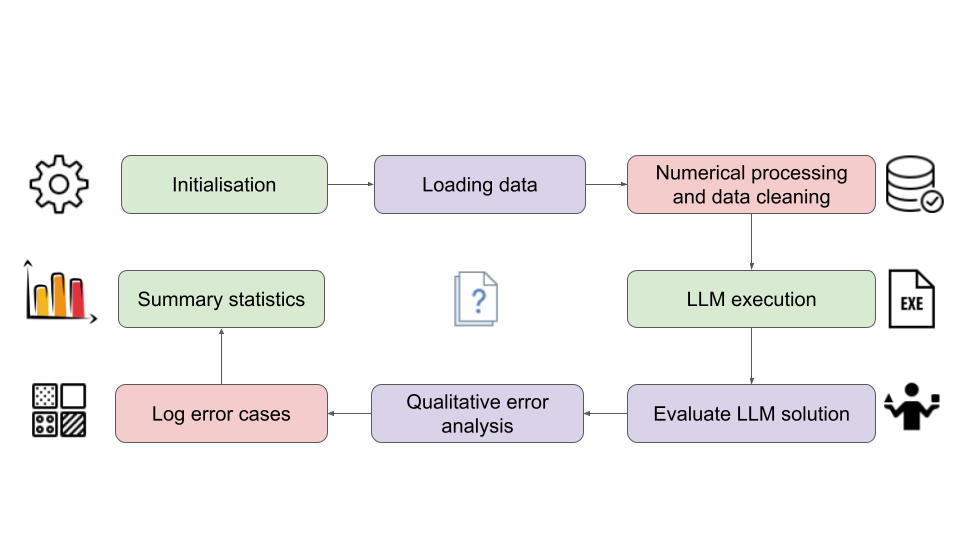}
    \caption{Workflow of evaluating LLMs for mathematical problem datasets.}
    \label{fig:datasets_details_workflow}
\end{figure*}

Our framework covers the whole process from initial setup, data loading, numerical processing, model execution, result comparison, to error recording and statistical output (Figure \ref{fig:datasets_details_workflow}). Each part is designed as an independent and interconnected unit, which is easy to expand and tune in different scenarios and provides an effective means for teaching, testing, model performance comparison and prompt tuning.


\begin{itemize}
  \item Step 1 - Initialisation: In the framework's initial stage, we should first complete the necessary configuration work to ensure that all parameters and modules are ready. Researchers need to specify the model name used (such as gpt-4o, gpt-3.5-turbo), then they should set the API Key\cite{tesfagiorgis2023large}, and specify the evaluation data file path (such as train.jsonl) and the number of samples (such as 500 questions). This part ensures the smoothness of data calls and model access in subsequent experimental processes.

  \item Step 2 - Loading Data: Our framework considers the problem (dataset) from a  JSONL \cite{escarda2024llms}   file to prepare the LLM input. We define each row as a JSONL object that contains the description of the math problem and the standard answer information \cite{schubotz2016semantification}. Parsing this JSONL data line by line builds a complete list of questions. It provides a standardised data source for subsequent model inputs, ensuring that each sample is processed and evaluated in the desired format \cite {gallen2024importance}.

  \item Step 3 - Numerical processing and data cleaning: We must perform uniform numerical treatment on the model output and the standard answer\cite{jiang2021can} to make the comparison of answers more accurate. The core operations of this stage include removing non-numeric characters from answers, formatting the data in the form of "1,200\%" and converting it into the standard number "1200"; converting integer form to string representation (for example, converting 1200.0 to "1200"). At the same time, we also need to use regular expressions to extract the last digit part of the answer. The primary purpose of this step is to eliminate comparison errors due to format differences, thus ensuring the accuracy of subsequent evaluation results\cite{macedo2024exploring}.
  
  \item Step 4 - LLM execution: After complete data pre-processing, the framework automatically sends a question request to the LLMs. The system prompt word is pre-set in the request to ask the model to return only a "pure number" answer without any explanation \cite{macedo2024exploring} to ensure that the LLMs output results are in a uniform format. This step not only effectively controls the content of the answer output so that it can only output what we want, but also allows the evaluation process to focus only on numerical comparison.  It reduces the risk of ambiguity in the model's understanding of the question \ cite{laskar2024systematic}.
  
  \item Step 5 - Evaluate LLM solution: 
 Our framework needs to clean the LLM output and the standard answer numerically, and then perform strict comparison in the form of a string matching \cite{frieder2023mathematical}. In this way, it can accurately judge whether the solution of the model is consistent with the standard answer, to obtain the correct result of each question, and calculate the accuracy rate\cite{chang2024survey}. This process is the core of assessing the mathematical ability of LLMs and provides direct data support for mathematicians \cite{spreitzer2024mathematical}.
 
 \item Step 6 - Qualitative Error Analysis: Based on the evaluation results, representative model outputs are manually analysed to identify typical errors in the reasoning chain. Typical errors in the reasoning chain are identified. This stage supplements the details that cannot be captured by the automated scoring and enhances the explanatory nature of the assessment.

  \item Step 7 - Log Error Cases: The evaluation system records all cases of errors \cite{pan2025lemma} to further analyse the model's performance on a specific topic. The record's content includes the question number,  the question, the standard answer and the output of the LLMs. Therefore, we  can further explore the causes of errors, and provide data and empirical support for model improvement and prompt optimisation
  
  \item Step 8 - Summary statistics  of   results: Following the above evaluations, we need to count and print the overall evaluation results. The system will compute the total accuracy based on the comparison results. 
  Meanwhile, the evaluation framework will also output the evaluation summary as a JSONL file or print it out directly. It is human-readable material and a reference for immediate judgment of model performance.
\end{itemize}

\subsection{Evaluation}

We utilise our LLMs evaluation framework to evaluate the performance of the GPT-4o, DeepSeek, and Gemini-2.0 models in providing accurate responses. The evaluation framework covers four main objectives to measure the accuracy and reliability of the model comprehensively.

\begin{itemize}

    \item Accuracy: This will help us determine whether the answer finally output by the LLM is consistent with the standard answer. It is the most direct indicator to measure the model's problem-solving capability. The accuracy of the solution has been commonly used as the primary evaluation criterion of LLMs \cite{rane2024gemini}.

    \item Reasoning: Refers to the logical clarity, rationality and degree of conformity to the semantics of the question of the intermediate reasoning steps generated by the model. Reasoning quality focuses on whether there are logical problems, such as skipping steps, circular reasoning, and unfounded assumptions \cite{xu2025large}.
    
    \item Completeness: Even if the final answer is correct, if the intermediate steps are missing or part of the reasoning is skipped, it is considered lacking in completeness. Therefore, \textit{completeness} ensures whether the LLM covers all the key steps required to solve the problem\cite{ashqar2025benchmarking}.
    
    \item Consistency: This enables us to evaluate if the LLM can ensure consistency in logical structure\cite{parmar2024logicbench}, and answer output while avoiding unstable reasoning through randomness or language fluctuation.
    
\end{itemize}

\section{Results}

We comprehensively evaluate the advantages and disadvantages of LLMs in mathematical problem-solving tasks by examining the key indicators, including accuracy, reasoning quality, completeness and consistency. Furthermore, the evaluation framework utilises the Chain of Thought \cite{jiang2024llms}, Few-sample and Zero-sample \cite{luo2024pkrd} techniques, and multi-round verification processes \cite{hoffreumon2021multi} to improve the accuracy of the evaluation results.

\subsection{LLMs for solving GSM8K problems}

\begin{table}[htbp]
\centering
\caption{ Accuracy (\%) of LLMs for GSM8K Datasets}
\label{tab:gsm8k}
\begin{tabular}{|l|c|c|c|}
\toprule
\textbf{Model} & \textbf{GPT-4o} & \textbf{DeepSeek} &\ \textbf{Gemini}\\
\midrule
\textbf{Accuracy} & 50.0\% & 42.6\% & \textbf{53.9\% }\\
\bottomrule
\end{tabular}
\end{table}

We systematically evaluate the selected LLMs based on the GSM8K dataset, denoting basic mathematical reasoning tasks, as this dataset comprises 5,000 questions. Although its difficulty level belongs to elementary mathematics, covering typical question types such as multi-step problems, basic arithmetic operations, and unit conversions, the questions are relatively simple regarding knowledge difficulty \cite{lai5002356solving}. However, this dataset poses a comprehensive challenge to LLMs' logical operations, organisation and computational capabilities.


We investigated more about the errors contributing to this mediocre result, including interruptions in the reasoning chain, omissions of steps, misunderstandings of quantitative relationships, and errors in unit conversion. In the case of GSM8K, all three models sometimes arrives at the correct answer. However, they cannot generate intermediate steps due to an unstable reasoning process or improper handling of details, which reflects their lack of coherence and robustness in multi-step tasks.

Although all three models can assist in solving basic maths tasks, they can still fail to consistently provide stable and plausible correct answers when faced with reasoning-based problems such as GSM8K. This suggests that the reasoning capabilities of LLMs cannot yet meet the practical needs for robustness and interpretability of fundamental mathematics in an educational setting.

\begin{table}[htbp]
\centering
\caption{Common reasoning errors in GSM8K (sample-based)}
\label{tab:gsm8k_errors}
\begin{tabular}{p{3cm}p{5cm}}
\toprule
\textbf{Error Type} & \textbf{Description}\\
\midrule
Step Skipping & Model omits key intermediate steps required to reach the final answer.\\
Unit Confusion & Incorrect unit conversions.\\
Misinterpretation & Misunderstand conception and comparative or conditional phrases.\\
Arithmetic Mistakes & All steps are correct but the final answer is wrong.\\
\bottomrule
\end{tabular}
\end{table}

Table \ref{tab:gsm8k} shows that Gemini performs relatively best on the GSM8K dataset, with an accuracy of 53.9\%. In addition, it demonstrates a more complete reasoning path compared to GPT-4o (50.0\%) and \textcolor{black}{DeepSeek-V3 (42.6\%) \footnote{\url{https://github.com/sydney-machine-learning/LLM-mathematics}}.}  To better understand this difference, we qualitatively (manually) review representative model responses, as summarised in Table \ref{tab:gsm8k_errors}. Gemini produced fewer instances of step skipping, unit confusion, and misinterpretation of the question intent. In contrast, we also found that DeepSeek's reasoning chain is the most unstable, with frequent errors in unit conversion and logical expression, and is at the bottom of the three in terms of stability and accuracy.

\subsection{LLMs for MATH500}
\begin{table}[htbp]
\centering
\caption{ Accuracy (\%) of LLMs for MATH500 Datasets}
\label{tab:math500}
\begin{tabular}{|l|c|c|c|}
\toprule
\textbf{Model} & \textbf{GPT-4o} & \textbf{DeepSeek} &\ \textbf{Gemini}\\
\midrule
\textbf{Accuracy} & 36.8\% & 51.1\% & \textbf{54.7\% } \\
\bottomrule
\end{tabular}
\end{table}

\begin{figure*}[htbp]
\centering
\includegraphics[width=\linewidth]{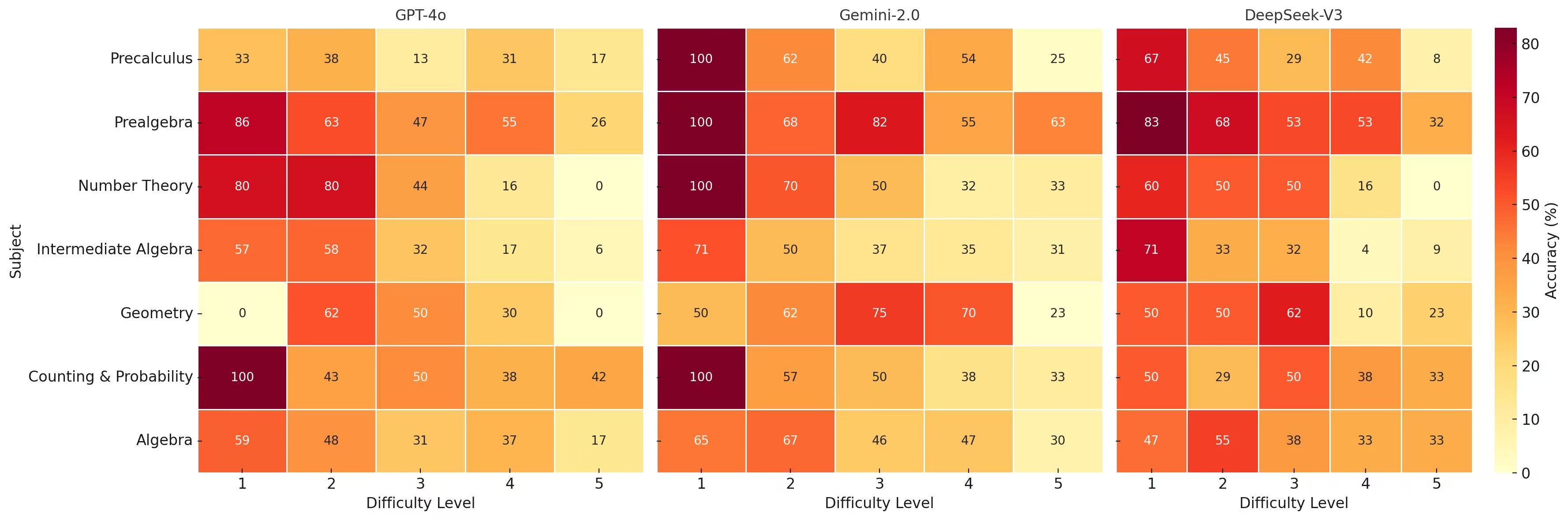}
\caption{Accuracy heatmaps of GPT-4o, Gemini-2.0, and DeepSeek-V3 on MATH500 by subject and difficulty level.}
\label{fig:math500_heatmaps}
\end{figure*}
In the case of MATH500 problems using LLMs, we evaluate the accuracy of the answers in terms of two dimensions: the difficulty level of the questions and the domain involved in the questions.  MATH500 consists of problems from seven subjects: Algebra, Counting and Probability, Geometry, Intermediate Algebra, Number Theory, Prealgebra, and Precalculus.  The issues in Math500 are more theoretical than those in GSM8K and are mainly used to test the application of knowledge and formulas. 
 We subdivided the difficulty level into five levels based on the length and characteristics of the words in the questions, and the specific division criteria are outlined in Table \ref{tab:level_definition}.

\begin{table*}[htbp]
    \centering
    \caption{MATH500 - difficulty level}
    \label{tab:level_definition}
    \renewcommand{\arraystretch}{1.4}
    \begin{tabular}{c m{8cm} m{2.5cm} m{4cm}}
    \toprule
    \textbf{Level}&\textbf{Features}&\textbf{Average Length}&\textbf{Sample}\\
    \midrule
    1 & Single-step arithmetic, explicit expression, no transformation required & 20-25 words & Solving quadratic equations, simple substitution\\
    2 & Multi-step but clearly structured, no hidden conditions & 25-30 words & Area calculations, unit conversion problems\\
    3 & Abstract expressions, multi-step reasoning, implicit relationships & 30-35 words & Quadratic properties, categorical reasoning\\
    4 & Multi-step reasoning, need for constructive method, non-intuitive structures & 35-40 words & Number theory analysis, path counting problems\\
    5 & Highly abstract, complex conditions, proof-like problem structures & >40 words & Conditional probabilistic chained derivations, function constructions\\
    \bottomrule
    \end{tabular}
\end{table*}



\begin{table*}[htbp]
\centering
\caption{Accuracy (\%) of LLMs on MATH500 by Difficulty Level and Subject}
\label{tab:math500_accuracy}
\begin{tabular}{m{2cm}m{5cm}ccc}
\toprule
  \textbf{Category} & \textbf{Item} & \textbf{DeepSeek} & \textbf{Gemini} & \textbf{GPT} \\
  \midrule
  \multirow{5}{*}{\textbf{Level}} 
  & Level 1 & 58\% & \textbf{79\%} & 62\% \\
  & Level 2 & 51\% & \textbf{66\%} & 63\% \\
  & Level 3 & 44\% & \textbf{61\%} & 45\% \\
  & Level 4 & 36\% & \textbf{53\%} & 44\% \\
  & Level 5 & 29\% & \textbf{45\%} & 26\% \\
\midrule
\multirow{7}{*}{\textbf{Subject}} 
  & Prealgebra & 52\% & \textbf{69\%} & 51\% \\
  & Count/Probability & 37\% & \textbf{45\%} & \textbf{45\%} \\
  & Algebra & 40\% & \textbf{49\%} & 35\% \\
  & Geometry & 34\% & \textbf{53\%} & 30\% \\
  & Number Theory & 31\% & \textbf{48\%} & 35\% \\
  & Precalculus & 30\% & \textbf{50\%} & 26\% \\
  & Intermediate Algebra & \textbf{38\%} & \textbf{38\%} & 25\% \\
\bottomrule
\end{tabular}
\end{table*}



We present the performance of the respective LLMs for the MATH500 dataset in Table \ref{tab:math500_accuracy}. Although it handles lower and intermediate difficulty problems with reasonable success, GPT-4o's accuracy declines significantly on higher-level tasks.  GPT-4o achieves an accuracy of 62\% on Level 1 and 63\% on Level 2 questions, comparable to DeepSeek but lower than Gemini’s performance (79\% and 66\%, respectively). These results indicate that GPT-4o can handle basic computation and well-structured problems with relatively stable reasoning.  Gemini 2.0 performs relatively consistently in questions of lower difficulty (level 1 and level 2), reaching an average accuracy of about 79\% and 65\%, respectively. However, when the difficulty level is raised to level 3 and above, the accuracy rate drops significantly and gradually decreases with increasing difficulty. Gemini 2.0 reports the lowest accuracy rate in level 5 questions, achieving only about 35\%. This trend reveals Gemini's weak ability to deal with complex mathematical tasks such as abstract algebra and combinatorial counting, and in particular, cannot master multi-step logical reasoning paths. We can see that the overall accuracy of DeepSeek is not high; level 1 has the highest accuracy rate of 62.8\%. However, as the difficulty increases, the accuracy rate rapidly decreases from level 2 and is all below 50\% and is less than 30\% afterwards. This indicates that DeepSeek does not perform well in handling some mathematical problems, especially those that are slightly more complex or difficult.

Table \ref{tab:math500_accuracy} demonstrates that Gemini performed relatively well in Prealgebra and Geometry, with accuracy rates of approximately 70\% and 54\%, respectively. However, Gemini's answer accuracy is the lowest at 38\% and 45\% on the Intermediate Algebra and Counting and Probability questions. This result further reveals that the model has better generalisation ability in simple questions with clear structure and rules. Still, it shows some limitations in compound questions involving implicit reasoning and conditional construction. Among the subjects covered by DeepSeek, only Prealgebra has an accuracy rate of exactly 50\%, while the other subjects have an accuracy rate of around 30\%. Among them, the worst-performing subject is Intermediate Algebra, with an accuracy rate of only 18.6\%, indicating that DeepSeek does not perform well in dealing with mathematical problems that do not have a complete logical framework and conditional limitations.



In summary, GPT-4o performs reasonably well on low to moderately difficult problems, but its accuracy drops significantly on complex issues requiring multi-step reasoning. Despite its excellent language generation capabilities, the model often suffers from incomplete reasoning chains, skipped steps, or logic breaks in real-world mathematical tasks, especially in scenarios where the response length or reasoning time is limited. In contrast, DeepSeek-V3 has the lowest overall accuracy, and its performance decreases the most as the difficulty of the questions increases. The model occasionally performs well on basic arithmetic or probability questions, but does not perform well on problems with dense algebraic structure or open logic. Its output often lacks key intermediate steps, and the reasoning process lacks coherence.  Gemini-2.0 demonstrates the most stable and scalable mathematical reasoning on the MATH500 dataset. GPT-4o achieves a balance between language and problem-solving ability in easy tasks. DeepSeek-V3 has a weaker overall performance and is mainly suited for low-difficulty tasks or scenarios that require tool-call support. 
In future evaluations, using Chain-of-Thought (CoT) prompting \cite{zhang2022automatic}can help by encouraging the model to show its step-by-step thinking. CoT can improve both performance and explainability, giving a more accurate picture of what the model can do.

To investigate the interaction effect of the model on different branches of mathematics and the difficulty of the questions, this study constructs a heat map of the accuracy of Gemini-2.0 on the MATH500 dataset based on the dimensions of "subject area" and "difficulty level", as shown in Figure \ref{fig:math500_heatmaps}. The heat map shows that questions in Precalculus and Number Theory are primarily distributed in the lower difficulty levels, with almost no problems in level 4 and level 5 involved. In contrast, problems in pre-algebra are widely distributed in levels 1 and 3, and some appear in level 2 and level 5, indicating that the accuracy in this area contains both basic introductory questions and more complex extensions. 
The heatmap (Figure \ref{fig:math500_heatmaps}) in the case of  DeepSeek on the MATH500 dataset shows that most of the questions of Precalculus, Prealgebra, Number Theory, and Intermediate Algebra have relatively low difficulty levels, mainly distributed at levels 1 and 2, with almost no high difficulty questions. In contrast, most Geometry problems are distributed around level 3, probably due to their higher complexity and the need for more constraints. Therefore, DeepSeek has a few advantages in solving simple problems, such as basic pre-algebra or counting and probability. Although DeepSeek performs averagely in other aspects, the high difficulty problems it solves demonstrate its potential for improvement.

\subsection{LLMs for MIT Open Courseware Dataset}


\begin{table}[htbp]
\centering
\caption{ \hl{Accuracy (\%) of LLMs for MIT Open Courseware Datasets}}
\label{tab:university}
\begin{tabular}{|l|c|c|c|}
\toprule
\textbf{Model} & \textbf{GPT-4o} & \textbf{DeepSeek} &\textbf{Gemini}\\
\midrule
\textbf{Accuracy} & 35.6\% & 42.8\%  & \textbf{54.4\%} \\
\bottomrule
\end{tabular}
\end{table}


\textcolor{black}{We evaluated LLM performance on the MIT Open Courseware dataset}, building on the qualitative analysis of incorrect answers using a structured scoring rubric as shown in Figure \ref{fig:LLMs_evaluation_workflow}.  We designed this rubric to assess the quality of mathematical reasoning, presentation, and accuracy with  criteria shown in Table \ref{tab:criteria}: 

\begin{table}[htbp]
    \centering
    \caption{The Score Criteria}
    \label{tab:criteria}
    \begin{tabular}{p{1cm} p{7cm}}
    \toprule
    \textbf{Score}&\textbf{Criteria}\\
    \midrule
     1 & A completely incorrect solution with major logical flaws or missing fundamental steps\\
     2 & A weak understanding with multiple errors or incoherent reasoning\\
     3 & The main method is correct, but contains explanation or calculation problems\\
     4 & A mostly correct solution with minor mistakes or slightly informal reasoning\\
     5 & Fully correct, well-organised, and rigorous answers that represent a model solution\\
    \bottomrule
    \end{tabular}
\end{table}

We excluded all sub-questions that received a full score of 5. For each non-perfect response, we conducted detailed error analysis to identify the primary cause of the mistake. We assigned each response to only one dominant error category to ensure consistency, even if multiple issues were present. This approach formed the foundation for our subsequent analysis of error types. We developed a taxonomy of ten categories to describe and classify the nature of the errors, which reflect different aspects of failure in LLM-generated reasoning  as shown in Table \ref{tab:ms_of_University}. 


\begin{table*}[htbp]
\centering
\small
\caption{\hl{LLM performance based on scores on University problems by subject, where higher scores indicate higher correctness of solutions.} }
\label{tab:scores_of_University}
\begin{tabular}{lccccc}
\toprule
\textbf{Subject} & \textbf{Score} & \textbf{DeepSeek} & \textbf{Gemini} & \textbf{GPT-4o} \\
\midrule
\multirow{5}{*}{\textbf{Optimisation}} 
  & 1 & \textbf{6.7\%} & 0.0\% & 0.0\% \\
  & 2 & \textbf{6.7\%} & 0.0\% & 0.0\% \\
  & 3 & \textbf{20.0\%} & 0.0\% & 0.0\% \\
  & 4 & 6.7\% & 20.0\% & \textbf{53.0\%} \\
  & 5 & 60\% & \textbf{80\%} & 46.7\% \\
\midrule
\multirow{5}{*}{\textbf{Computational Methods for Finance}} 
  & 1 & \textbf{19.0\%} & 2.6\% & 0.0\% \\
  & 2 & \textbf{10.1\%} & 6.5\% & 1.2\% \\
  & 3 & 10.1\% & \textbf{31.2\%} & 8.8\% \\
  & 4 & 10.1\% & 40.3\% & \textbf{42.5\%} \\
  & 5 & \textbf{50.6\%} & 19.5\% & 47.5\% \\
\midrule
\multirow{5}{*}{\textbf{Statistical Inference}} 
  & 1 & \textbf{33.3\%} & 12.8\% & 0.0\% \\
  & 2 & \textbf{35.9\%} & 25.8\% & 0.0\% \\
  & 3 & 7.7\% & \textbf{17.9\%} & 7.7\% \\
  & 4 & 5.1\% & 15.4\% & \textbf{33.3\%} \\
  & 5 & 17.9\% & 28.2\% & \textbf{59.0\%} \\
\bottomrule
\end{tabular}
\end{table*}


\textcolor{black}{Table \ref{tab:scores_of_University} shows that the GPT-4o score distribution differs significantly between subjects in the University dataset.} In Statistical Inference, the model achieves the highest proportion of perfect scores (59.0\%). This indicates strong performance in problems involving standard statistical procedures such as MLE (Maximum Likelihood Estimation). This suggests that GPT-4o is well-suited to problems with familiar mathematical structures and clear solution paths. In optimisation, GPT-4o has the highest proportion of 4-point responses (53.3\%), with slightly fewer perfect scores (46.7\%). This may reflect that while GPT-4o can follow correct procedures, such as computing gradients, it often lacks full formal justification and presentation, which prevents it from achieving a perfect score under strict grading criteria. For Computational Methods for Finance, the model has the highest rate of low scores (3 and below, totalling around 10\%). These errors may stem from difficulties interpreting domain-specific terminology, probabilistic reasoning, or less familiar symbolic conventions. Overall, GPT-4o’s performance depends heavily on problem type, with stronger results on structurally straightforward and procedurally familiar tasks.

\begin{table*}[htbp]
    \centering
    \caption{Mainly Shortcoming Details}
    \label{tab:shortcoming}
    \begin{tabular}{lp{10cm}}
    \toprule
    \textbf{Type}&\textbf{Description}\\
    \midrule
    Understanding Errors & Misunderstanding of key concepts, such as confusion between prior, posterior, and MLE\\
    Deduction Errors & Correct understanding of concepts, but errors in formula derivation or transformation steps\\
    Expression Errors & Correct calculations or reasoning, but unclear writing, lack of explanation, or confusing symbols\\
    Process Control Errors& Disorganised problem-solving process, inconsistent answers to sub-questions, or missing conditions\\
    Format Errors & Minor errors that do not affect the core logic, such as copying formulas incorrectly, writing variables incorrectly, or missing symbols\\
    \bottomrule
    \end{tabular}
\end{table*}

 The SCoT framework facilitates a clearer evaluation of LLM performance by decomposing their reasoning into structured steps. The average score of GPT-4o performs well in Statistical Inference course (see Table \ref{tab:university_avg_score}) mainly because it follows standard statistical procedures and generates coherent solutions that align with the SCoT criteria, particularly in terms of logical completeness and consistency. This matches the high proportion of perfect scores in Statistical Inference, showing that GPT-4o performs well when the problem is clear and the reasoning is well-defined.
We designed a prompt that guides GPT-4o through solving each sub-question step-by-step, and an automated grading model based on Deepseek-V3 that evaluates each solution step. The LLM returns a structured JSON format answer for each step which allows us to check the logic and correctness at every stage, check the model's reasoning, and easily spot gaps. This step-by-step evaluation ensures that we can detect issues in the model's reasoning, such as missing intermediate steps, inaccurate calculations, or insufficient explanations.Table \ref{tab:ms_of_University} shows the types of errors made by GPT-4o on the advanced mathematical problems. In addition, we also reviewed and categorised the errors manually from the model’s results that did not receive full marks, helping to highlight common mistakes in multi-step mathematical problem solving with the following:
\begin{itemize}
    \item Lack of Justification or Explanation (39\%)
    \item Less Precise (33\%)
    \item Missing Intermediate Steps or Details (9\%)
\end{itemize}

\begin{table}[htbp]
\centering
\small
\caption{Main shortcomings of answers on University problems dataset}
\label{tab:ms_of_University}
\begin{tabular}{lcc}
\toprule
\textbf{Error type} & \textbf{Model} & \textbf{Percent}\\
\midrule
\multirow{3}{*}{Lack of justification or explanation} 
  & GPT & \textbf{39\%} \\
  & DeepSeek & 21\% \\
  & Gemini & 20\% \\
\midrule
\multirow{3}{*}{Less precise}
  & GPT & \textbf{33\% }\\
  & DeepSeek& 3\% \\
  & Gemini & 3\% \\
\midrule
\multirow{3}{*}{Missing intermediate steps or details}
  & GPT & 9\% \\
  & DeepSeek & \textbf{24\%} \\
  & Gemini & 18\% \\
\midrule
\multirow{3}{*}{Unclear or in consistent notation}
  & GPT & 8\% \\
  & DeepSeek & 7\% \\
  & Gemini & \textbf{18\%} \\
\midrule
\multirow{3}{*}{Poor structure or lack of formal presentation}
  & GPT & 3\% \\
  & DeepSeek & 14\% \\
  & Gemini & \textbf{17\%} \\
\midrule
\multirow{3}{*}{Incomplete gradient/Hessian analysis}
  & GPT & \textbf{3\% }\\
  & DeepSeek & \textbf{3\% } \\
  & Gemini & \textbf{3\% } \\
\midrule
\multirow{3}{*}{Lack of logical flow}
  & GPT & 3\% \\
  & DeepSeek & \textbf{9\% }\\
  & Gemini & 2\% \\
\midrule
\multirow{3}{*}{Missing eigenvalue/matrix-related reasoning}
  & GPT & 2\% \\
  & DeepSeek & - \\
  & Gemini & \textbf{7\%} \\
\midrule
\multirow{3}{*}{Verbose}
  & GPT & - \\
  & DeepSeek & 6\% \\
  & Gemini & \textbf{9\%} \\
\bottomrule
\end{tabular}
\end{table}

These three types of errors account for over 80\% of all observed errors(see Table \ref{tab:ms_of_University}), which indicates that GPT-4o can often identify the correct approach or final answer, but it frequently omits essential reasoning or fails to maintain computational precision. At the same time, the remaining error types related to notation clarity, logical coherence, and incomplete second-order analysis in optimisation problems which were observed less frequent but still noteworthy, \textcolor{black}{This findings are based on manual review of model responses in the University dataset, as summarised in Table \ref{tab:ms_of_University}}.

In general, most of GPT-4o's errors (Table \ref{tab:ms_of_University}) are mainly due to insufficient formal expressions rather than a lack of understanding of mathematical concepts. Therefore, introducing stronger logical constraints and formal specifications in the prompt design is expected to substantially improve the rigour of model reasoning and problem-solving completeness. These findings suggest that improving the model’s ability to intermediate reasoning and clarity of its logical progression may significantly improve GPT-4o's overall performance on mathematical tasks.

In the case of DeekSeek (Table \ref{tab:scores_of_University}), the distribution of scores is quite extreme, mainly distributed in Score 1, Score 2, and Score 5, with fewer middle scores. This indicates that DeepSeek either answers the questions perfectly or has many errors, and there are fewer incomplete answers. The reason for this phenomenon may be that DeepSeek specialises in specific fields, or it may be due to the underlying logic of the code itself that led to this result. Furthermore, we can observe that DeepSeek performs much better in computational methods and optimisation than in statistical inference. This also indicates that DeepSeek has an advantage in handling computational and optimisation problems, but is slightly inferior in analysis, statistics, and explanation, and can only solve relatively simple problems. Furthermore, by analysing the main shortcomings of answers, as shown in Table \ref{tab:ms_of_University}, it can be found that the most common shortcomings that DeepSeek encounters when answering are missing intermediate steps or details and a lack of justification or explanation. This is mainly manifested in DeepSeek providing answers with only partial methods, explanations, and conclusions, but lacking intermediate processes and detailed computational details. However, this may be related to the difficulty of the question, as relatively simple questions usually have a more detailed process, while more difficult questions may omit a large number of steps.

 The distribution of scores shows that Gemini performs well in multi-step calculations and flexible application of specific formulas, Table \ref{tab:scores_of_University}. The scores of answers in solving problems related to statistical inference are very close, which indicates that a mix of formulas and analysis can be a challenge for Gemini. Issues like these require solvers to tackle subsequent questions based on the results you derived from the former ones. Even if Gemini can retain all the context, it still cannot realise and correct previous mistakes in solving the rest of the problems. The problem-solving process lacks self-censorship; hence, the more subproblems there are, the higher the probability of getting a low score.

The shortcomings of Gemini-2.0, as shown in Table \ref{tab:ms_of_University} highlight verbosity as a significant drawback where Gemini’s answers often take a longer path to the destination compared with the correct solution. Strangely, verbosity and lack of explicit explanation of key theory or variables can appear in an answer at the same time. Students know what they have learned in this course and try to use the knowledge to complete the exam. However, Gemini knows almost everything about this topic from a wide range of sources and it cannot figure out which key formula or concept is being examined in this problem \footnote{\url{https://github.com/sydney-machine-learning/LLM-mathematics}}. Gemini omits some main steps in the process of proof for the same reason. It takes some computational steps and interpretation for granted. However, such omissions can lead to a loss of marks when the answer is marked.

As an LLM, Gemini prefers to use natural language instead of mathematical language to explain conditions or variables. This is also why the answer is verbose, although using plain language is helpful for beginners to understand the core idea of this problem. The marker may get confused about the solutions expressed and lower the grade due to weak presentation, since Gemini chooses the notation freely and arbitrarily. Even though notation can vary across texts in different reference books, there is still a consensus on the usage of notation among students and scholars. Non-standard and uncommon notation may mislead the solver to a wrong result. The formatting issues mentioned above are not important if the final result is correct; however, it will make the answers ambiguous and less precise.

  As shown in Table \ref{tab:ms_of_University}, the problems requiring proof and calculations are mainly included in the dataset across three fields, Optimisation, Computational Methods for Finance and Statistical Inference. Subjects selected are close to industrial practices, since the performance of LLMs has to be validated through real-world applications rather than pure mathematical theory. 29\% accuracy is not much surprising due to the rigorous criterion. Only a full mark answer can be considered correct for proof problems. 

\subsection{Qualitative assessment for University problems}

\textcolor{black}{We studied Statistical Inference LLM solutions from the University problems dataset qualitatively to evaluate their capability in dealing with complex mathematical inference questions.} The question consists of four sub-problems, which examine the determination of sufficiency and completeness, the unbiased verification of maximum likelihood estimation(MLE), the derivation of asymptotic distributions, and the application of multivariate delta methods, and involves several core concepts in statistical theory, with a rigorous logical structure and a coherent step-by-step approach, which makes it an ideal sample for evaluating the performance of LLMs to perform mathematical reasoning \footnote{\url{https://github.com/sydney-machine-learning/LLM-mathematics}}. 
The results of the study, as shown in the Table \ref{tab:summary_university_stat}, show that the mainstream models usually identify basic solution strategies and can initially construct a reasoning framework. However, the models generally have problems in the specific derivation process. We take the completeness proof to provide an example. Although the model can identify sufficient statistics, it often fails to apply theoretical tools such as the Lehmann-Scheffé theorem; in the unbiasedness test, although part of the model can correctly judge the direction, it lacks the detailed development of the expectation calculation; in the application of the multivariate delta method, even though it is close to being correct in form, it lacks the matrix and parameter independence premise. Although the form is nearly correct in applying the multivariate Delta method, there is a lack of a matrix inversion step and parameter independence. These problems can be categorised as 'reasoning chain breaks' or 'logical jumps', especially in the case of a longer structure and higher logical relevance of sub-problems. The lack of rigour in the early reasoning of the model will be cumulatively amplified to the subsequent sub-problems, resulting in the so-called 'cumulative collapse'. The so-called 'cumulative breakdown' occurs.
In addition, dimensional analyses of the SCoT framework show that its output struggles to meet the basic requirements of logical closure and mathematical rigour for higher-order reasoning problems. These errors are also repeated to varying degrees in other university questions, reflecting that there are still some stability and controllability challenges in formal mathematical tasks in the current LLM \footnote{\url{https://github.com/sydney-machine-learning/LLM-mathematics}}.

\begin{table*}[htbp]
\centering
\caption{Summary of LLM's performance on a four-part statistical inference problem}
\label{tab:summary_university_stat}
\begin{tabular}{cp{3.5cm}p{5cm}p{3cm}}
\toprule
\textbf{Sub-question} & \textbf{Content} & \textbf{LLMs Output Summary} & \textbf{Common Issue} \\
\midrule
Q1 & Sufficiency and completeness justification & Identified exponential family structure correctly, but failed to invoke formal completeness arguments such as Lehmann–Scheffé theorem. & Missing statistical justification \\
\midrule
Q2 & Unbiasedness of MLE for $1/(1+\theta)$ & Provided correct final answer but omitted expected value calculation and skipped integration steps. & Incomplete derivation \\
\midrule
Q3 & Asymptotic distribution (Delta Method) & Applied Delta Method without deriving the transformation derivative or specifying Fisher Information. & Lack of precision and justification \\
\midrule
Q4 & Multivariate Delta Method application & General form used correctly, but failed to mention parameter independence or invert the Fisher Information matrix. & Incomplete assumptions and structure \\
\bottomrule
\end{tabular}

\vspace{0.5cm}
\begin{minipage}{\linewidth}
\footnotesize
\textit{Note: The above summary is based on current models' outputs. Further analysis will incorporate additional model results for a more comprehensive comparison.}
\end{minipage}
\end{table*}

\subsection{Summary of results: LLMs across the three mathematical problem datasets}

\textcolor{black}{We finally present a summary of results to review the strengths and weaknesses of the respective LLMs across the three mathematical problem datasets, including GSM8K, MATH500 and University problem datasets.  }

As shown in Table \ref{tab:overall_accuracy}, the results show that the accuracy of  GPT-4o, DeepSeek, and Gemini for the GSM8K dataset is 54.7\%, 54.4\%, and 53.9\% , respectively. This is generally maintained at a medium-low level from 50\% to 55\%, and the difference between the models is small, among which  GPT-4o has a slight advantage. 
 GSM8K contains problems that require the LLMs to find the results step by step, which requires logical reasoning and simple calculation. GSM8K problems can be solved easily by college students and even high school students \cite{cobbe2021training}. However, we find that the performance of LLMs on GSM8K tasks is average, and it seems that there is a  50 \% chance of getting it right.  The similar accuracy indicates that the problem does not lie in the difference between models, but in the limitation of logical reasoning within LLMs themselves, which makes them not suitable for this set of problems. Therefore, all three models can be helpful when facing problems like GSM8K, but they cannot guarantee correct answers.

 The overall accuracy for University problems is 29\%, which is satisfactory in Table \ref{tab:overall_accuracy}. The main reason for this accuracy is that a majority of answers receive scores of 3 or 4, especially problems from Computational Methods for Finance.
DeepSeek has a much higher accuracy rate in handling University problems (unlike MATH500), with an overall accuracy rate of 42.8\%, Table \ref{tab:overall_accuracy}.


\textcolor{black}{Finally, we further summarise the overall performance of LLMs under different subjects in the University dataset by computing the weighted average score of each LLM under each subject based on the distribution of scores in Table \ref{tab:scores_of_University}.} We compute the average score $S$ by:
\begin{equation}
S=\sum_{i=1}^{5} i \cdot p_i
\end{equation}
where \(p_i\) denotes the proportion of the model receiving score i in that subject. As shown in Table \ref{tab:university_avg_score}, GPT-4o has a stable average score in several domains, especially in University  \textit{Statistical Inference}, where it performs the best, with a score of 4.51. This reflecting its good adaptability to standard statistical tasks. Gemini shows a clear advantage in the \textit{Optimisation} problem, with an average score of 4.80, but it still suffers from some redundancy in linguistic expression and notation specification. In contrast, the performance of DeepSeek-V3 varies greatly across subjects, with a fair score of 3.63 in \textit{Computational Methods for Finance}, but a significantly weaker score of 2.38 in \textit{Statistical Inference}, mainly due to an incomplete inference chain and missing intermediate steps. These weighted average scores provide quantitative support for differences in the capabilities of LLMs in domain-specific tasks and further corroborate what the previous section revealed based on the quality of the reasoning chain.

\begin{table*}[htbp!]
\centering
\caption{\hl{Average Score of LLMs for University Dataset by Subject}}
\renewcommand{\arraystretch}{1.5}
\begin{tabular}{lccc}
\toprule
\textbf{Subject}&\textbf{DeepSeek}&\textbf{Gemini}&\textbf{GPT-4o}\\
\midrule
Optimisation & 4.07 & \textbf{4.80} & 4.46\\
Computational Methods for Finance & 3.63 & 3.68 & \textbf{4.36}\\
Statistical Inference & 2.38 & 3.21 & \textbf{4.51}\\
\bottomrule
\end{tabular}
\label{tab:university_avg_score}
\end{table*}

\begin{table*}[htbp]
\centering
\caption{\hl{Overall Accuracy (\%) of LLMs on Three Different Datasets}}
\label{tab:overall_accuracy}
\begin{tabular}{m{3cm}m{7cm}m{2cm}}
\toprule
\textbf{Model} & \textbf{Dataset} & \textbf{Accuracy}\\
\midrule
\multirow{3}{*}{\textbf{DeepSeek}} 
  & GSM8K & 42.6\% \\
  & MATH500 & \textbf{51.1\%} \\
  & University & 42.8\% \\
\midrule
\multirow{3}{*}{\textbf{Gemini-2.0}} 
  & GSM8K & 53.9\%  \\
  & MATH500 & \textbf{54.7\% } \\
  & University& 54.4\%  \\
\midrule
\multirow{3}{*}{\textbf{GPT-4o}} 
  & GSM8K & \textbf{50.0\%} \\
  & MATH500 & 36.8\% \\
  & University& 35.6\% \\
\bottomrule
\end{tabular}
\end{table*}




We compared three LLMs, including GPT-4o, Gemini-2.0, and DeepSeek-V3—on mathematical problem-solving across varying difficulty levels. \textcolor{black}{In addition to benchmark datasets, we used the University dataset to assess models that feature diverse problem formats and non-standardised expressions, requiring both procedural precision and semantic comprehension.} We observed that the LLMs can stably output complete answers containing intermediate variables and formula calculation steps when faced with moderately complex algebra, geometry, and probability questions. Furthermore, we wanted to evaluate whether the LLMs can automatically decompose sub-tasks and determine solution strategies, with relatively clear reasoning logic. Gemini-2.0 exhibited the most stable and structured reasoning patterns, particularly excelling in multi-step tasks through its chain-of-thought generation. GPT-4o shows high final-answer accuracy on straightforward problems but frequently omits intermediate steps and struggles with issues involving implicit constraints. In contrast, DeepSeek-V3 can reconstruct the multi-step problem-solving process more stably and performs well in problems centred on concept understanding. However, its effectiveness is significantly reduced when dealing with issues involving ambiguous language or implicit phrasing.
Although each LLM demonstrated particular areas of strength, we also observed that they all have certain limitations in interpreting questions accurately, adapting reasoning flexibly, and maintaining logical completeness. These findings show that the reasonable design of prompt words and the addition of intermediate-step guidance are crucial to improving the robustness of the model.


\begin{table*}[htbp]
\centering
\caption{Effect of Prompt Design on LLMs Responses (Illustrative Cases)}
\label{tab:prompt_effect}
\begin{tabular}{p{5cm}p{1.5cm}p{2.5cm}p{5.5cm}}
\toprule
\textbf{Prompt Type} & \textbf{Model} & \textbf{Outcome} & \textbf{Statement}\\
\midrule
Find the total cost & GPT-4o & incorrect & Model failed to perform unit conversion and no intermediate steps shown.\\
Show all steps to calculate the total cost and unit conversions & GPT-4o & correct & Model gave all clear steps and accurate calculations.\\
\midrule
Solve the value of X & DeepSeek & incorrect & Skipped variable substitution and used wrong formula.\\
State the formula and justify each step to find X & DeepSeek & correct & Using the correct formula and giving the logical justification for each step.\\
\midrule
Solve this problem & Gemini & Partially correct & Answer included but reasoning was vaggue and generalised.\\
Show the each step for solving this problem and give the statement & Gemini & correct & Provided well-structured reasoning with consistent logic.\\
\bottomrule
\end{tabular}
\end{table*}

\begin{table}[htbp]
\centering
\caption{Qualitative Comparison of LLMs on the Univesrity Dataset}
\label{tab:qual_university_summary}
\begin{tabular}{lp{1.2cm}p{1.2cm}p{1.2cm}}
\toprule
\textbf{Dimension} & \textbf{GPT-4o} & \textbf{DeepSeek} & \textbf{Gemini}\\
\midrule
Final answer accuracy & High & Moderate & Moderate\\
Step completeness & Moderate & Low & High\\
Step validity & High & Moderate & High\\
Intermediate calculation & Moderate & High & High\\
Problem understanding & Moderate & Low & High\\
Handing of complex logic & High & Low & Moderate\\
\bottomrule
\end{tabular}
\end{table}

\textcolor{black}{The results indicate that GPT-4o has an acceptable level of performance when responding to the University questions (Table \ref{tab:scores_of_University}).} Specifically, it shows a high final-answer accuracy across simple and moderately difficult problems, achieving an average correctness rate of 82\%. However, a detailed analysis of the SCoT framework reveals multiple problems. The reasoning process tends to "skip steps." Though the ultimate answer could be right, the logical coherence and consistency of the middle steps are lost. This is especially crucial for the case of applied problems because GPT-4o does not tend to include the implicit conditions set by the question, leading to low problem understanding and computational correctness scores, resulting in a low step completeness score of 57\%  (Table \ref{tab:ms_of_University}). In addition, it is less likely to have explicit equations and relations between the variables, hurting the interpretability of the solution approach. Notably, GPT-4o’s performance is sensitive to the formulation of the prompt. \textcolor{black}{We observe notable improvements in the model's logical coherence and reasoning completeness when prompts explicitly request step-by-step instructions or clarify assumptions, particularly on complex University problems (see Table \ref{tab:qual_university_summary} and Table \ref{tab:prompt_effect}).} Although the model can produce correct results, its reliability in educational contexts depends heavily on structured prompt scaffolding.

As shown in Tables \ref{tab:ms_of_University} and \ref{tab:qual_university_summary},  our qualitative analysis for the University dataset shows that DeepSeek-V3 performs well on conceptual and definitional questions with a computational accuracy of 81\% and step validity rate of 76\%, which is superior to that of GPT-4o, and can reconstruct the mathematical solution chain. However, the question comprehension score drops to 64\% when faced with problems with high linguistic ambiguity, making it challenging to identify implied structural changes or unit conversions in the stem. The model is more dependent on system prompts, and complex problems may lead to correct answers but incorrect reasoning, with the risk of misleading instructions. Gemini-2.0 has a balanced performance on all five dimensions, with a step completeness rate of 82\% and an intermediate accuracy of 80 \% (Table \ref{tab:ms_of_University}), and can generate a stable reasoning process. It handles complex problems well, maintaining semantic consistency and logical coherence, which leads to a drop in final accuracy to 66\%.


\textcolor{black}{Through an in-depth analysis of University higher-level mathematical problems, we found that the design of prompts significantly impacts the reasoning performance of  LLMs in higher education scenarios. The questions in the University dataset require open-mindness, an academic style of expression, and step-by-step solutions, which challenge the model's linguistic comprehension and logical reasoning ability.} We observe that there are differences in the way different models respond to the prompts. GPT-4o significantly improves the completeness of its reasoning chain when explicitly asked to answer in steps. Gemini maintains a high level of linguistic consistency and clarity of expression in response to structured questions. Finally, DeepDeek is more prone to logical jumps or erroneous results in response to a lack of derivation steps.

Recent studies have focused on evaluating the efficiency of LLMs in mathematical reasoning tasks and providing insights by constructing new benchmark tests. Mirzadeh et al.  \cite{mirzadeh2024gsm} and Deng et al.\cite{deng2024explicit} revealed the limitations of these models in mathematical reasoning, and the potential advantages suggested by Chain-of-Thought (CoT) hint at potential advantages. The results of our research match these findings and apply the SCoT framework to model evaluation, focusing on multiple performance dimensions. In addition, this study explores the model's adaptability in an educational setting by introducing University higher-level course topics, providing insights for pedagogical applications.

\textcolor{black}{This study has practical implications for teaching mathematics at colleges and universities, and also applies to the rest of Australian universities.} In traditional lectures and tutorials, the focus of assessment grading is qualitative analysis of the rigour of students' reasoning in breaking two the problem and providing steps of how the solution has been reached. On the contrary, with well-designed hints, LLMs can automatically generate more complete solutions and provide hints for different types of errors (e.g., missing steps, unclear definitions, and computational errors), which can help improve the efficiency of instructional feedback. \textcolor{black}{For example, for statistical inference or optimisation questions, the Lecturer can set up a standard prompt template to guide the model in producing a structured solution path in JSON format for students' self-checking and assessment.} The SCoT framework used in this study can be extended to the classroom as a teaching assessment tool. For example, 'completeness of steps', 'accuracy of intermediate calculations', and 'problem comprehension' can be used as scoring dimensions to assess students' submitted assignments or model-generated process answers. This way, a 'human-machine co-assessment' mechanism can be constructed to achieve a more transparent and detailed teaching feedback system.

In future studies, we suggest integrating LLM into the teaching platform to build an intelligent teaching and learning system with subject suitability and role division of labour. Educators (lecturers/tutors/teachers) can preset the prompting strategies and assessment criteria for different course modules(e.g., probabilistic inference, matrix decomposition, financial modelling), and the system will call on various models to complete the explanation, training and correction tasks. This nested design, based on 'course structure + model characteristics + prompting strategies', is expected to enhance the application value of LLM in advanced mathematics courses and promote the two-way improvement of personalised learning and teaching efficiency.

\section{Conclusion}

 This study comprehensively evaluated the performance of GPT-4o, DeepSeek-V3, and Gemini-2.0 in mathematical problem-solving tasks in five dimensions, covering three types of datasets, including basic (GSM8K), intermediate (MATH500), and advanced (University). This study presents an automatic scoring framework to evaluate LLMs on mathematical problems featuring multidimensional qualitative analysis. It not only reveals the performance differences between the current mainstream models in mathematical reasoning tasks but also verifies the key role of cue design and structured scoring mechanisms in improving the interpretability of the models.

 The results show that GPT-4o has strong stability in complex tasks, especially in statistical inference and optimisation problems. Gemini has a complete chain of reasoning and clear expression, which is advantageous in structured issues. DeepSeek has gone well in computational problem solving, but suffers from logic breaks and step omissions when dealing with problems with ambiguous language or many reasoning steps and missing steps.

 Although all three models are suitable for basic arithmetic and algebraic calculation, advanced and graduate-level problems still challenge their completeness of reasoning and logical coherence. This study offers empirical evidence of LLMs' mathematical reasoning capacity based on automated grading, human evaluation, and visual examination. It gives theoretical and practical support for the choice of models in an educational context and offers possibilities for future research in multi-modal tasks, prompt strategy adaptation, and human-reinforced learning. The results of this study provide theoretical support and practical paths for the selection of models, the construction of assisted teaching systems, and the optimisation of multi-step prompting strategies in subsequent educational scenarios.

\section*{Code and Data Availability}
\textcolor{black}{Data and code: }\href{ https://github.com/sydney-machine-learning/LLM-mathematics}{ https://github.com/sydney-machine-learning/LLM-mathematics}


\section{Author Contributions }

 Ruonan Wang contributed to writing (original draft) and editing, SCoT framework, and analysis. Runxi Wang contributed to Gemini API coding and analysis, the construction of University dataset, analysis, writing and editing. Chengfeng Wu contributed to DeepSeek API coding, the construction of the University dataset,  analysis, visualisation, and editing.  Qinglin Zhou contributed to GPT-4o API coding,  analysis, visualisation and  design of the experiments.  Yuwen Shen contributed to writing and editing,  analysis and results. 

R. Chandra contributed to conceptualisation, project supervision, editing, and analysis.

\nocite{*}
\bibliographystyle{IEEEtran}

\bibliography{refs}

\begin{thebibliography}{100}
\providecommand{\url}[1]{#1}
\csname url@samestyle\endcsname
\providecommand{\newblock}{\relax}
\providecommand{\bibinfo}[2]{#2}
\providecommand{\BIBentrySTDinterwordspacing}{\spaceskip=0pt\relax}
\providecommand{\BIBentryALTinterwordstretchfactor}{4}
\providecommand{\BIBentryALTinterwordspacing}{\spaceskip=\fontdimen2\font plus
\BIBentryALTinterwordstretchfactor\fontdimen3\font minus \fontdimen4\font\relax}
\providecommand{\BIBforeignlanguage}[2]{{%
\expandafter\ifx\csname l@#1\endcsname\relax
\typeout{** WARNING: IEEEtran.bst: No hyphenation pattern has been}%
\typeout{** loaded for the language `#1'. Using the pattern for}%
\typeout{** the default language instead.}%
\else
\language=\csname l@#1\endcsname
\fi
#2}}
\providecommand{\BIBdecl}{\relax}
\BIBdecl

\bibitem{liljedahl2019mathematical}
P.~Liljedahl and M.~Santos-Trigo, \emph{Mathematical problem solving}.\hskip 1em plus 0.5em minus 0.4em\relax Springer, 2019.

\bibitem{tambychik2010students}
T.~Tambychik and T.~S.~M. Meerah, ``Students' difficulties in mathematics problem-solving: What do they say?'' \emph{Procedia-Social and Behavioral Sciences}, vol.~8, pp. 142--151, 2010.

\bibitem{mitchelmore2004abstraction}
M.~Mitchelmore and P.~White, ``Abstraction in mathematics and mathematics learning.'' \emph{International Group for the Psychology of Mathematics Education}, 2004.

\bibitem{kinard2008rigorous}
J.~T. Kinard and A.~Kozulin, \emph{Rigorous mathematical thinking: Conceptual formation in the mathematics classroom}.\hskip 1em plus 0.5em minus 0.4em\relax Cambridge University Press, 2008.

\bibitem{yushau2004mathematics}
B.~Yushau, M.~Bokhari, A.~Mji, and D.~Wessels, ``Mathematics: conceptions, learning and teaching,'' \emph{King Fahd University of Petroleum \& Minerals, Department of Mathematical Sciences: Technical Report Series: TR}, vol. 322, 2004.

\bibitem{hembree1990nature}
R.~Hembree, ``The nature, effects, and relief of mathematics anxiety,'' \emph{Journal for research in mathematics education}, vol.~21, no.~1, pp. 33--46, 1990.

\bibitem{susac2014development}
A.~Susac, A.~Bubic, A.~Vrbanc, and M.~Planinic, ``Development of abstract mathematical reasoning: the case of algebra,'' \emph{Frontiers in Human Neuroscience}, vol.~8, p. 679, 2014.

\bibitem{ab2019logical}
J.~Ab, G.~Margono, and W.~Rahayu, ``The logical thinking ability: Mathematical disposition and self-regulated learning,'' in \emph{Journal of Physics: Conference Series}, vol. 1155, no.~1.\hskip 1em plus 0.5em minus 0.4em\relax IOP Publishing, 2019, p. 012092.

\bibitem{firdaus2015developing}
F.~Firdaus, I.~Kailani, M.~N.~B. Bakar, and B.~Bakry, ``Developing critical thinking skills of students in mathematics learning,'' \emph{Journal of Education and Learning (EduLearn)}, vol.~9, no.~3, pp. 226--236, 2015.

\bibitem{peter2012critical}
E.~E. Peter, ``Critical thinking: Essence for teaching mathematics and mathematics problem solving skills,'' \emph{African Journal of Mathematics and Computer Science Research}, vol.~5, no.~3, pp. 39--43, 2012.

\bibitem{bin2022artificial}
M.~Z. bin Mohamed, R.~Hidayat, N.~N. binti Suhaizi, M.~K.~H. bin Mahmud, S.~N. binti Baharuddin \emph{et~al.}, ``Artificial intelligence in mathematics education: A systematic literature review,'' \emph{International Electronic Journal of Mathematics Education}, vol.~17, no.~3, p. em0694, 2022.

\bibitem{lazar2015importance}
S.~Lazar, ``The importance of educational technology in teaching,'' \emph{International Journal of Cognitive Research in Science, Engineering and Education}, vol.~3, no.~1, pp. 111--114, 2015.

\bibitem{voskoglou2020benefits}
M.~G. Voskoglou and A.-B.~M. Salem, ``Benefits and limitations of the artificial with respect to the traditional learning of mathematics,'' \emph{Mathematics}, vol.~8, no.~4, p. 611, 2020.

\bibitem{telegina2017use}
N.~V. Telegina, E.~G. Galimova, and S.~G. Dobrotvorskaya, ``The use of interactive learning technologies in math classes,'' \emph{Modern journal of language teaching methods}, vol.~7, no.~4, pp. 106--113, 2017.

\bibitem{larreamendy2006going}
J.~Larreamendy-Joerns and G.~Leinhardt, ``Going the distance with online education,'' \emph{Review of educational research}, vol.~76, no.~4, pp. 567--605, 2006.

\bibitem{zubiaga2024natural}
A.~Zubiaga, ``Natural language processing in the era of large language models,'' p. 1350306, 2024.

\bibitem{gao2021introduction}
P.~Gao, J.~Li, and S.~Liu, ``An introduction to key technology in artificial intelligence and big data driven e-learning and e-education,'' \emph{Mobile Networks and Applications}, vol.~26, no.~5, pp. 2123--2126, 2021.

\bibitem{matzakos2023learning}
N.~Matzakos, S.~Doukakis, and M.~Moundridou, ``Learning mathematics with large language models: A comparative study with computer algebra systems and other tools,'' \emph{International Journal of Emerging Technologies in Learning (iJET)}, vol.~18, no.~20, pp. 51--71, 2023.

\bibitem{johnsen2024large}
M.~Johnsen, \emph{Large language models (LLMs)}.\hskip 1em plus 0.5em minus 0.4em\relax Maria Johnsen, 2024.

\bibitem{shanahan2024talking}
M.~Shanahan, ``Talking about large language models,'' \emph{Communications of the ACM}, vol.~67, no.~2, pp. 68--79, 2024.

\bibitem{hadi2023survey}
M.~U. Hadi, R.~Qureshi, A.~Shah, M.~Irfan, A.~Zafar, M.~B. Shaikh, N.~Akhtar, J.~Wu, S.~Mirjalili \emph{et~al.}, ``A survey on large language models: Applications, challenges, limitations, and practical usage,'' \emph{Authorea Preprints}, vol.~3, 2023.

\bibitem{zhao2023survey}
W.~X. Zhao, K.~Zhou, J.~Li, T.~Tang, X.~Wang, Y.~Hou, Y.~Min, B.~Zhang, J.~Zhang, Z.~Dong \emph{et~al.}, ``A survey of large language models,'' \emph{arXiv preprint arXiv:2303.18223}, vol.~1, no.~2, 2023.

\bibitem{kaddour2023challenges}
J.~Kaddour, J.~Harris, M.~Mozes, H.~Bradley, R.~Raileanu, and R.~McHardy, ``Challenges and applications of large language models,'' \emph{arXiv preprint arXiv:2307.10169}, 2023.

\bibitem{vemuri2024evolution}
V.~Vemuri, ``The evolution of human-computer interaction: From command lines to conversational interfaces powered by large language models,'' \emph{J Artif Intell Mach Learn \& Data Sci 2024}, vol.~2, no.~1, pp. 2257--2266.

\bibitem{kothari2024enhancing}
D.~K. Kothari and O.~N.~N. Fernando, ``Enhancing human-computer interaction through ai: A study on chatgpt in educational environments,'' in \emph{2024 IEEE Conference on Artificial Intelligence (CAI)}.\hskip 1em plus 0.5em minus 0.4em\relax IEEE, 2024, pp. 500--503.

\bibitem{hadi2023large}
M.~U. Hadi, R.~Qureshi, A.~Shah, M.~Irfan, A.~Zafar, M.~B. Shaikh, N.~Akhtar, J.~Wu, S.~Mirjalili \emph{et~al.}, ``Large language models: a comprehensive survey of its applications, challenges, limitations, and future prospects,'' \emph{Authorea Preprints}, vol.~1, pp. 1--26, 2023.

\bibitem{laleh2024survey}
A.~R. Laleh and M.~N. Ahmadabadi, ``A survey on enhancing reinforcement learning in complex environments: Insights from human and llm feedback,'' \emph{arXiv preprint arXiv:2411.13410}, 2024.

\bibitem{puerta2025multifaceted}
M.~Puerta-Beldarrain, O.~G{\'o}mez-Carmona, R.~S{\'a}nchez-Corcuera, D.~Casado-Mansilla, D.~L{\'o}pez-de Ipi{\~n}a, and L.~Chen, ``A multifaceted vision of the human-ai collaboration: a comprehensive review,'' \emph{IEEE Access}, 2025.

\bibitem{yan2024practical}
L.~Yan, L.~Sha, L.~Zhao, Y.~Li, R.~Martinez-Maldonado, G.~Chen, X.~Li, Y.~Jin, and D.~Ga{\v{s}}evi{\'c}, ``Practical and ethical challenges of large language models in education: A systematic scoping review,'' \emph{British Journal of Educational Technology}, vol.~55, no.~1, pp. 90--112, 2024.

\bibitem{luan2020challenges}
H.~Luan, P.~Geczy, H.~Lai, J.~Gobert, S.~J. Yang, H.~Ogata, J.~Baltes, R.~Guerra, P.~Li, and C.-C. Tsai, ``Challenges and future directions of big data and artificial intelligence in education,'' \emph{Frontiers in psychology}, vol.~11, p. 580820, 2020.

\bibitem{karlstrom2024exploring}
K.~Karlstr{\"o}m, ``Exploring the possibility and implementation of ai-supported online math coaching,'' 2024.

\bibitem{opesemowo2024artificial}
O.~A. Opesemowo and M.~Ndlovu, ``Artificial intelligence in mathematics education: The good, the bad, and the ugly,'' \emph{Journal of Pedagogical Research}, vol.~8, no.~3, pp. 333--346, 2024.

\bibitem{tan2022information}
J.~Tan, ``Information analysis of advanced mathematics education-adaptive algorithm based on big data,'' \emph{Mathematical Problems in Engineering}, vol. 2022, no.~1, p. 7796681, 2022.

\bibitem{lecun2015deep}
Y.~LeCun, Y.~Bengio, and G.~Hinton, ``Deep learning,'' \emph{nature}, vol. 521, no. 7553, pp. 436--444, 2015.

\bibitem{wu2018development}
Y.-c. Wu and J.-w. Feng, ``Development and application of artificial neural network,'' \emph{Wireless Personal Communications}, vol. 102, pp. 1645--1656, 2018.

\bibitem{kumar2024large}
P.~Kumar, ``Large language models (llms): survey, technical frameworks, and future challenges,'' \emph{Artificial Intelligence Review}, vol.~57, no.~10, p. 260, 2024.

\bibitem{chen2024large}
J.~Chen, Z.~Liu, X.~Huang, C.~Wu, Q.~Liu, G.~Jiang, Y.~Pu, Y.~Lei, X.~Chen, X.~Wang \emph{et~al.}, ``When large language models meet personalization: Perspectives of challenges and opportunities,'' \emph{World Wide Web}, vol.~27, no.~4, p.~42, 2024.

\bibitem{jiang2024survey}
J.~Jiang, F.~Wang, J.~Shen, S.~Kim, and S.~Kim, ``A survey on large language models for code generation,'' \emph{arXiv preprint arXiv:2406.00515}, 2024.

\bibitem{liu2024datasets}
Y.~Liu, J.~Cao, C.~Liu, K.~Ding, and L.~Jin, ``Datasets for large language models: A comprehensive survey,'' \emph{arXiv preprint arXiv:2402.18041}, 2024.

\bibitem{meshkin2024harnessing}
H.~Meshkin, J.~Zirkle, G.~Arabidarrehdor, A.~Chaturbedi, S.~Chakravartula, J.~Mann, B.~Thrasher, and Z.~Li, ``Harnessing large language models’ zero-shot and few-shot learning capabilities for regulatory research,'' \emph{Briefings in Bioinformatics}, vol.~25, no.~5, p. bbae354, 2024.

\bibitem{alto2023modern}
V.~Alto, \emph{Modern Generative AI with ChatGPT and OpenAI Models: Leverage the capabilities of OpenAI's LLM for productivity and innovation with GPT3 and GPT4}.\hskip 1em plus 0.5em minus 0.4em\relax Packt Publishing Ltd, 2023.

\bibitem{imran2024google}
M.~Imran and N.~Almusharraf, ``Google gemini as a next generation ai educational tool: a review of emerging educational technology,'' \emph{Smart Learning Environments}, vol.~11, no.~1, p.~22, 2024.

\bibitem{neha2025survey}
F.~Neha and D.~Bhati, ``A survey of deepseek models,'' \emph{Authorea Preprints}, 2025.

\bibitem{rahman2025comparative}
A.~Rahman, S.~H. Mahir, M.~T.~A. Tashrif, A.~A. Aishi, M.~A. Karim, D.~Kundu, T.~Debnath, M.~A.~A. Moududi, and M.~Eidmum, ``Comparative analysis based on deepseek, chatgpt, and google gemini: Features, techniques, performance, future prospects,'' \emph{arXiv preprint arXiv:2503.04783}, 2025.

\bibitem{gallegos2024bias}
I.~O. Gallegos, R.~A. Rossi, J.~Barrow, M.~M. Tanjim, S.~Kim, F.~Dernoncourt, T.~Yu, R.~Zhang, and N.~K. Ahmed, ``Bias and fairness in large language models: A survey,'' \emph{Computational Linguistics}, vol.~50, no.~3, pp. 1097--1179, 2024.

\bibitem{thapa2025large}
S.~Thapa, S.~Shiwakoti, S.~B. Shah, S.~Adhikari, H.~Veeramani, M.~Nasim, and U.~Naseem, ``Large language models (llm) in computational social science: prospects, current state, and challenges,'' \emph{Social Network Analysis and Mining}, vol.~15, no.~1, pp. 1--30, 2025.

\bibitem{myers2024foundation}
D.~Myers, R.~Mohawesh, V.~I. Chellaboina, A.~L. Sathvik, P.~Venkatesh, Y.-H. Ho, H.~Henshaw, M.~Alhawawreh, D.~Berdik, and Y.~Jararweh, ``Foundation and large language models: fundamentals, challenges, opportunities, and social impacts,'' \emph{Cluster Computing}, vol.~27, no.~1, pp. 1--26, 2024.

\bibitem{kamath2024llm}
U.~Kamath, K.~Keenan, G.~Somers, and S.~Sorenson, ``Llm challenges and solutions,'' in \emph{Large Language Models: A Deep Dive: Bridging Theory and Practice}.\hskip 1em plus 0.5em minus 0.4em\relax Springer, 2024, pp. 219--274.

\bibitem{hosseinpour2018step}
S.~Hosseinpour, M.~M.~R. Alavi~Milani, and H.~Pehlivan, ``A step-by-step solution methodology for mathematical expressions,'' \emph{Symmetry}, vol.~10, no.~7, p. 285, 2018.

\bibitem{lai5002356solving}
H.~Lai, B.~Wang, J.~Liu, F.~He, C.~Zhang, H.~Liu, and H.~Chen, ``Solving mathematical problems using large language models: A survey,'' \emph{Available at SSRN 5002356}.

\bibitem{davis2023testing}
E.~Davis and S.~Aaronson, ``Testing gpt-4 with wolfram alpha and code interpreter plug-ins on math and science problems,'' \emph{arXiv preprint arXiv:2308.05713}, 2023.

\bibitem{li2024assessing}
S.~Li, Y.~Cheng, J.~Chen, J.~Xuan, S.~He, and W.~Shang, ``Assessing the performance of ai-generated code: A case study on github copilot,'' in \emph{2024 IEEE 35th International Symposium on Software Reliability Engineering (ISSRE)}.\hskip 1em plus 0.5em minus 0.4em\relax IEEE, 2024, pp. 216--227.

\bibitem{bhattacharya2023exploring}
P.~Bhattacharya, M.~Chakraborty, K.~N. Palepu, V.~Pandey, I.~Dindorkar, R.~Rajpurohit, and R.~Gupta, ``Exploring large language models for code explanation,'' \emph{arXiv preprint arXiv:2310.16673}, 2023.

\bibitem{lyu2024automatic}
M.~R. Lyu, B.~Ray, A.~Roychoudhury, S.~H. Tan, and P.~Thongtanunam, ``Automatic programming: Large language models and beyond,'' \emph{ACM Transactions on Software Engineering and Methodology}, 2024.

\bibitem{zhu2023intelligent}
S.~Zhu, T.~Yu, T.~Xu, H.~Chen, S.~Dustdar, S.~Gigan, D.~Gunduz, E.~Hossain, Y.~Jin, F.~Lin \emph{et~al.}, ``Intelligent computing: the latest advances, challenges, and future,'' \emph{Intelligent Computing}, vol.~2, p. 0006, 2023.

\bibitem{dai2019bridging}
W.-Z. Dai, Q.~Xu, Y.~Yu, and Z.-H. Zhou, ``Bridging machine learning and logical reasoning by abductive learning,'' \emph{Advances in Neural Information Processing Systems}, vol.~32, 2019.

\bibitem{liu2024mathbench}
H.~Liu, Z.~Zheng, Y.~Qiao, H.~Duan, Z.~Fei, F.~Zhou, W.~Zhang, S.~Zhang, D.~Lin, and K.~Chen, ``Mathbench: Evaluating the theory and application proficiency of llms with a hierarchical mathematics benchmark,'' \emph{arXiv preprint arXiv:2405.12209}, 2024.

\bibitem{didolkar2024metacognitive}
A.~Didolkar, A.~Goyal, N.~R. Ke, S.~Guo, M.~Valko, T.~Lillicrap, D.~Jimenez~Rezende, Y.~Bengio, M.~C. Mozer, and S.~Arora, ``Metacognitive capabilities of llms: An exploration in mathematical problem solving,'' \emph{Advances in Neural Information Processing Systems}, vol.~37, pp. 19\,783--19\,812, 2024.

\bibitem{laskar2023systematic}
M.~T.~R. Laskar, M.~S. Bari, M.~Rahman, M.~A.~H. Bhuiyan, S.~Joty, and J.~X. Huang, ``A systematic study and comprehensive evaluation of chatgpt on benchmark datasets,'' \emph{arXiv preprint arXiv:2305.18486}, 2023.

\bibitem{chang2024survey}
Y.~Chang, X.~Wang, J.~Wang, Y.~Wu, L.~Yang, K.~Zhu, H.~Chen, X.~Yi, C.~Wang, Y.~Wang \emph{et~al.}, ``A survey on evaluation of large language models,'' \emph{ACM transactions on intelligent systems and technology}, vol.~15, no.~3, pp. 1--45, 2024.

\bibitem{surampudi2024big}
Y.~Surampudi, \emph{Big Data Meets LLMs: A New Era of Incident Monitoring}.\hskip 1em plus 0.5em minus 0.4em\relax Libertatem Media Private Limited, 2024.

\bibitem{castillo2023effect}
A.~G.~R. Castillo, G.~S. Silva, J.~F. Arocutipa, H.~Q. Berrios, M.~A.~M. Rodriguez, G.~Y. Reyes, H.~R.~P. Lopez, R.~M.~V. Teves, H.~V.~H. Rivera, and J.~L. Arias-Gonz{\'a}les, ``Effect of chat gpt on the digitized learning process of university students,'' \emph{Journal of Namibian Studies: History Politics Culture}, vol.~33, no.~1, pp. 1--15, 2023.

\bibitem{zhai2024effects}
C.~Zhai, S.~Wibowo, and L.~D. Li, ``The effects of over-reliance on ai dialogue systems on students' cognitive abilities: a systematic review,'' \emph{Smart Learning Environments}, vol.~11, no.~1, p.~28, 2024.

\bibitem{susnjak2024chatgpt}
T.~Susnjak and T.~R. McIntosh, ``Chatgpt: The end of online exam integrity?'' \emph{Education Sciences}, vol.~14, no.~6, p. 656, 2024.

\bibitem{nguyen2023ethical}
A.~Nguyen, H.~N. Ngo, Y.~Hong, B.~Dang, and B.-P.~T. Nguyen, ``Ethical principles for artificial intelligence in education,'' \emph{Education and information technologies}, vol.~28, no.~4, pp. 4221--4241, 2023.

\bibitem{yang2024formal}
K.~Yang, G.~Poesia, J.~He, W.~Li, K.~Lauter, S.~Chaudhuri, and D.~Song, ``Formal mathematical reasoning: A new frontier in ai,'' \emph{arXiv preprint arXiv:2412.16075}, 2024.

\bibitem{monib2024generative}
W.~K. Monib, A.~Qazi, R.~A. Apong, M.~T. Azizan, L.~De~Silva, and H.~Yassin, ``Generative ai and future education: a review, theoretical validation, and authors’ perspective on challenges and solutions,'' \emph{PeerJ Computer Science}, vol.~10, p. e2105, 2024.

\bibitem{motlagh2023impact}
N.~Y. Motlagh, M.~Khajavi, A.~Sharifi, and M.~Ahmadi, ``The impact of artificial intelligence on the evolution of digital education: A comparative study of openai text generation tools including chatgpt, bing chat, bard, and ernie,'' \emph{arXiv preprint arXiv:2309.02029}, 2023.

\bibitem{janiesch2021machine}
C.~Janiesch, P.~Zschech, and K.~Heinrich, ``Machine learning and deep learning,'' \emph{Electronic markets}, vol.~31, no.~3, pp. 685--695, 2021.

\bibitem{rusk2016deep}
N.~Rusk, ``Deep learning,'' \emph{Nature Methods}, vol.~13, no.~1, pp. 35--35, 2016.

\bibitem{wang2024scaling}
C.~Wang, Y.~Wang, Z.~Li, Q.~Jia, and W.~Liu, ``Scaling laws of data-driven machine learning models: A survey and taxonomy,'' \emph{Authorea Preprints}, 2024.

\bibitem{thota43deep}
N.~P.~R. Thota and A.~Lakshmanarao, ``Deep learning \& its applications,'' \emph{INDEX, VOLUME I}, vol.~43, p.~81.

\bibitem{hao2016deep}
X.~Hao, G.~Zhang, and S.~Ma, ``Deep learning,'' \emph{International Journal of Semantic Computing}, vol.~10, no.~03, pp. 417--439, 2016.

\bibitem{mehrotra2019basics}
D.~Mehrotra, \emph{Basics of artificial intelligence \& machine learning}.\hskip 1em plus 0.5em minus 0.4em\relax Notion Press, 2019.

\bibitem{singh2023study}
S.~Singh and S.~Hooda, ``A study of challenges and limitations to applying machine learning to highly unstructured data,'' in \emph{2023 7th International Conference On Computing, Communication, Control And Automation (ICCUBEA)}.\hskip 1em plus 0.5em minus 0.4em\relax IEEE, 2023, pp. 1--6.

\bibitem{liu2011supervised}
B.~Liu, ``Supervised learning,'' in \emph{Web Data Mining: Exploring Hyperlinks, Contents, and Usage Data}.\hskip 1em plus 0.5em minus 0.4em\relax Springer, 2011, pp. 63--132.

\bibitem{luo2023self}
Q.~Luo, W.~Zeng, M.~Chen, G.~Peng, X.~Yuan, and Q.~Yin, ``Self-attention and transformers: Driving the evolution of large language models,'' in \emph{2023 IEEE 6th International Conference on Electronic Information and Communication Technology (ICEICT)}.\hskip 1em plus 0.5em minus 0.4em\relax IEEE, 2023, pp. 401--405.

\bibitem{raiaan2024review}
M.~A.~K. Raiaan, M.~S.~H. Mukta, K.~Fatema, N.~M. Fahad, S.~Sakib, M.~M.~J. Mim, J.~Ahmad, M.~E. Ali, and S.~Azam, ``A review on large language models: Architectures, applications, taxonomies, open issues and challenges,'' \emph{IEEE access}, vol.~12, pp. 26\,839--26\,874, 2024.

\bibitem{juhasz2024large}
L.~Z. Juh{\'a}sz, ``Large language models, fine tuning, code generation,'' in \emph{2024 IEEE 7th International Conference and Workshop {\'O}buda on Electrical and Power Engineering (CANDO-EPE)}.\hskip 1em plus 0.5em minus 0.4em\relax IEEE, 2024, pp. 000\,191--000\,200.

\bibitem{rane2024machine}
N.~L. Rane, M.~Paramesha, S.~P. Choudhary, and J.~Rane, ``Machine learning and deep learning for big data analytics: A review of methods and applications,'' \emph{Partners Universal International Innovation Journal}, vol.~2, no.~3, pp. 172--197, 2024.

\bibitem{roumeliotis2023chatgpt}
K.~I. Roumeliotis and N.~D. Tselikas, ``Chatgpt and open-ai models: A preliminary review,'' \emph{Future Internet}, vol.~15, no.~6, p. 192, 2023.

\bibitem{yenduri2024gpt}
G.~Yenduri, M.~Ramalingam, G.~C. Selvi, Y.~Supriya, G.~Srivastava, P.~K.~R. Maddikunta, G.~D. Raj, R.~H. Jhaveri, B.~Prabadevi, W.~Wang \emph{et~al.}, ``Gpt (generative pre-trained transformer)--a comprehensive review on enabling technologies, potential applications, emerging challenges, and future directions,'' \emph{IEEE Access}, 2024.

\bibitem{wang2022research}
X.~Wang, Y.~Chen, W.~Liu, and W.~Tai, ``Research on text classification model based on self-attention mechanism and multi-neural network.'' in \emph{ICBASE}, 2022, pp. 244--256.

\bibitem{wang2024chatgpt}
T.~Wang and Q.~Zhu, ``Chatgpt--technical research model, capability analysis, and application prospects,'' in \emph{2024 IEEE 7th Advanced Information Technology, Electronic and Automation Control Conference (IAEAC)}, vol.~7.\hskip 1em plus 0.5em minus 0.4em\relax IEEE, 2024, pp. 787--796.

\bibitem{kaufmann2023survey}
T.~Kaufmann, P.~Weng, V.~Bengs, and E.~H{\"u}llermeier, ``A survey of reinforcement learning from human feedback,'' \emph{arXiv preprint arXiv:2312.14925}, vol.~10, 2023.

\bibitem{liu2023summary}
Y.~Liu, T.~Han, S.~Ma, J.~Zhang, Y.~Yang, J.~Tian, H.~He, A.~Li, M.~He, Z.~Liu \emph{et~al.}, ``Summary of chatgpt-related research and perspective towards the future of large language models,'' \emph{Meta-radiology}, vol.~1, no.~2, p. 100017, 2023.

\bibitem{liu2023external}
A.~Liu, ``External reasoning: Towards multi-large-language-models interchangeable assistance with human feedback,'' \emph{arXiv preprint arXiv:2307.12057}, 2023.

\bibitem{daher2024use}
W.~Daher and F.~Gierdien, ``Use of language by generative ai tools in mathematical problem solving: The case of chatgpt,'' \emph{African Journal of Research in Mathematics, Science and Technology Education}, vol.~28, no.~2, pp. 222--235, 2024.

\bibitem{dao2023investigating}
X.-Q. Dao and N.-B. Le, ``Investigating the effectiveness of chatgpt in mathematical reasoning and problem solving: Evidence from the vietnamese national high school graduation examination,'' \emph{arXiv preprint arXiv:2306.06331}, 2023.

\bibitem{sanchez2023chatgpt}
L.~M. S{\'a}nchez-Ruiz, S.~Moll-L{\'o}pez, A.~Nu{\~n}ez-P{\'e}rez, J.~A. Mora{\~n}o-Fern{\'a}ndez, and E.~Vega-Fleitas, ``Chatgpt challenges blended learning methodologies in engineering education: A case study in mathematics,'' \emph{Applied Sciences}, vol.~13, no.~10, p. 6039, 2023.

\bibitem{frieder2023mathematical}
S.~Frieder, L.~Pinchetti, R.-R. Griffiths, T.~Salvatori, T.~Lukasiewicz, P.~Petersen, and J.~Berner, ``Mathematical capabilities of chatgpt,'' \emph{Advances in neural information processing systems}, vol.~36, pp. 27\,699--27\,744, 2023.

\bibitem{youvan2025deepseek}
D.~C. Youvan, ``Deepseek and china's ai renaissance: The rise of the four dragons and six tigers,'' 2025.

\bibitem{liao2025deepseek}
H.~Liao, ``Deepseek large-scale model: technical analysis and development prospect,'' \emph{Journal of Computer Science and Electrical Engineering}, vol.~7, no.~1, pp. 33--37, 2025.

\bibitem{puspitasarideepseek}
F.~D. Puspitasari, C.~Zhang, S.~K. Dam, M.~Zhang, T.-H. Kim, C.~S. Hong, S.-H. Bae, C.~Qin, J.~Wei, G.~Wang \emph{et~al.}, ``Deepseek models: A comprehensive survey of methods and applications.''

\bibitem{he2025survey}
T.~He, H.~Li, J.~Chen, R.~Liu, Y.~Cao, L.~Liao, Z.~Zheng, Z.~Chu, J.~Liang, M.~Liu \emph{et~al.}, ``A survey on complex reasoning of large language models through the lens of self-evolution,'' 2025.

\bibitem{gan2025mixture}
W.~Gan, Z.~Ning, Z.~Qi, and P.~S. Yu, ``Mixture of experts (moe): A big data perspective,'' \emph{arXiv preprint arXiv:2501.16352}, 2025.

\bibitem{tie2025survey}
G.~Tie, Z.~Zhao, D.~Song, F.~Wei, R.~Zhou, Y.~Dai, W.~Yin, Z.~Yang, J.~Yan, Y.~Su \emph{et~al.}, ``A survey on post-training of large language models,'' \emph{arXiv preprint arXiv:2503.06072}, 2025.

\bibitem{miao2024ai}
Q.~Miao and F.-Y. Wang, ``Ai for mathematics,'' in \emph{Artificial Intelligence for Science (AI4S) Frontiers and Perspectives Based on Parallel Intelligence}.\hskip 1em plus 0.5em minus 0.4em\relax Springer, 2024, pp. 21--39.

\bibitem{wang2025dynamic}
L.~Wang, ``Dynamic chain-of-thought: Towards adaptive deep reasoning,'' \emph{arXiv preprint arXiv:2502.10428}, 2025.

\bibitem{shao2024deepseekmath}
Z.~Shao, P.~Wang, Q.~Zhu, R.~Xu, J.~Song, X.~Bi, H.~Zhang, M.~Zhang, Y.~Li, Y.~Wu \emph{et~al.}, ``Deepseekmath: Pushing the limits of mathematical reasoning in open language models,'' \emph{arXiv preprint arXiv:2402.03300}, 2024.

\bibitem{park2024ensembling}
S.~Park, X.~Liu, Y.~Gong, and E.~Choi, ``Ensembling large language models with process reward-guided tree search for better complex reasoning,'' \emph{arXiv preprint arXiv:2412.15797}, 2024.

\bibitem{liu2024deepseek}
A.~Liu, B.~Feng, B.~Xue, B.~Wang, B.~Wu, C.~Lu, C.~Zhao, C.~Deng, C.~Zhang, C.~Ruan \emph{et~al.}, ``Deepseek-v3 technical report,'' \emph{arXiv preprint arXiv:2412.19437}, 2024.

\bibitem{wang2025review}
C.~Wang and M.~Kantarcioglu, ``A review of deepseek models' key innovative techniques,'' \emph{arXiv preprint arXiv:2503.11486}, 2025.

\bibitem{guo2024deepseek}
D.~Guo, Q.~Zhu, D.~Yang, Z.~Xie, K.~Dong, W.~Zhang, G.~Chen, X.~Bi, Y.~Wu, Y.~Li \emph{et~al.}, ``Deepseek-coder: When the large language model meets programming--the rise of code intelligence,'' \emph{arXiv preprint arXiv:2401.14196}, 2024.

\bibitem{saeidnia2023welcome}
H.~R. Saeidnia, ``Welcome to the gemini era: Google deepmind and the information industry,'' \emph{Library Hi Tech News}, no. ahead-of-print, 2023.

\bibitem{rane2024gemini}
N.~Rane, S.~Choudhary, and J.~Rane, ``Gemini versus chatgpt: applications, performance, architecture, capabilities, and implementation,'' \emph{Journal of Applied Artificial Intelligence}, vol.~5, no.~1, pp. 69--93, 2024.

\bibitem{xiang2025towards}
V.~Xiang, C.~Snell, K.~Gandhi, A.~Albalak, A.~Singh, C.~Blagden, D.~Phung, R.~Rafailov, N.~Lile, D.~Mahan \emph{et~al.}, ``Towards system 2 reasoning in llms: Learning how to think with meta chain-of-though,'' \emph{arXiv preprint arXiv:2501.04682}, 2025.

\bibitem{campesato2024google}
O.~Campesato, \emph{Google Gemini for Python: Coding with Bard}.\hskip 1em plus 0.5em minus 0.4em\relax Stylus Publishing, LLC, 2024.

\bibitem{zhang2025or}
B.~Zhang and P.~Luo, ``Or-llm-agent: Automating modeling and solving of operations research optimization problem with reasoning large language model,'' \emph{arXiv preprint arXiv:2503.10009}, 2025.

\bibitem{ono2024evaluating}
K.~Ono and A.~Morita, ``Evaluating large language models: Chatgpt-4, mistral 8x7b, and google gemini benchmarked against mmlu,'' \emph{Authorea Preprints}, 2024.

\bibitem{sprague2024cot}
Z.~Sprague, F.~Yin, J.~D. Rodriguez, D.~Jiang, M.~Wadhwa, P.~Singhal, X.~Zhao, X.~Ye, K.~Mahowald, and G.~Durrett, ``To cot or not to cot? chain-of-thought helps mainly on math and symbolic reasoning,'' \emph{arXiv preprint arXiv:2409.12183}, 2024.

\bibitem{lewkowycz2022solving}
A.~Lewkowycz, A.~Andreassen, D.~Dohan, E.~Dyer, H.~Michalewski, V.~Ramasesh, A.~Slone, C.~Anil, I.~Schlag, T.~Gutman-Solo \emph{et~al.}, ``Solving quantitative reasoning problems with language models,'' \emph{Advances in Neural Information Processing Systems}, vol.~35, pp. 3843--3857, 2022.

\bibitem{cobbe2021training}
K.~Cobbe, V.~Kosaraju, M.~Bavarian, M.~Chen, H.~Jun, L.~Kaiser, M.~Plappert, J.~Tworek, J.~Hilton, R.~Nakano \emph{et~al.}, ``Training verifiers to solve math word problems,'' \emph{arXiv preprint arXiv:2110.14168}, 2021.

\bibitem{hendrycks2021measuring}
D.~Hendrycks, C.~Burns, S.~Kadavath, A.~Arora, S.~Basart, E.~Tang, D.~Song, and J.~Steinhardt, ``Measuring mathematical problem solving with the math dataset,'' \emph{arXiv preprint arXiv:2103.03874}, 2021.

\bibitem{lee2024applying}
G.-G. Lee, E.~Latif, X.~Wu, N.~Liu, and X.~Zhai, ``Applying large language models and chain-of-thought for automatic scoring,'' \emph{Computers and Education: Artificial Intelligence}, vol.~6, p. 100213, 2024.

\bibitem{varastehnezhad2024llm}
A.~VarastehNezhad, R.~Tavasoli, M.~Masumi, and F.~Taghiyareh, ``Llm performance assessment in computer science graduate entrance exams,'' in \emph{2024 11th International Symposium on Telecommunications (IST)}.\hskip 1em plus 0.5em minus 0.4em\relax IEEE, 2024, pp. 232--237.

\bibitem{kamoi2024evaluating}
R.~Kamoi, S.~S.~S. Das, R.~Lou, J.~J. Ahn, Y.~Zhao, X.~Lu, N.~Zhang, Y.~Zhang, R.~H. Zhang, S.~R. Vummanthala \emph{et~al.}, ``Evaluating llms at detecting errors in llm responses,'' \emph{arXiv preprint arXiv:2404.03602}, 2024.

\bibitem{onwuegbuzie2014exemplar}
A.~J. Onwuegbuzie and V.~T. Byers, ``An exemplar for combining the collection, analysis, and interpretations of verbal and nonverbal data in qualitative research,'' \emph{International Journal of Education}, vol.~6, no.~1, p. 183, 2014.

\bibitem{gupta2025beyond}
A.~Gupta, J.~Reddig, T.~Calo, D.~Weitekamp, and C.~J. MacLellan, ``Beyond final answers: Evaluating large language models for math tutoring,'' \emph{arXiv preprint arXiv:2503.16460}, 2025.

\bibitem{rane2023enhancing}
N.~Rane, ``Enhancing mathematical capabilities through chatgpt and similar generative artificial intelligence: Roles and challenges in solving mathematical problems,'' \emph{Available at SSRN 4603237}, 2023.

\bibitem{feng2024numerical}
G.~Feng, K.~Yang, Y.~Gu, X.~Ai, S.~Luo, J.~Sun, D.~He, Z.~Li, and L.~Wang, ``How numerical precision affects mathematical reasoning capabilities of llms,'' \emph{arXiv preprint arXiv:2410.13857}, 2024.

\bibitem{lu2024proof}
M.~Lu, B.~Delaware, and T.~Zhang, ``Proof automation with large language models,'' in \emph{Proceedings of the 39th IEEE/ACM International Conference on Automated Software Engineering}, 2024, pp. 1509--1520.

\bibitem{tesfagiorgis2023large}
Y.~G. Tesfagiorgis and B.~M. Monteiro~Silva, ``Large language models as an interface to interact with api tools in natural language,'' 2023.

\bibitem{escarda2024llms}
M.~Escarda-Fern{\'a}ndez, I.~L{\'o}pez-Riob{\'o}o-Botana, S.~Barro-Tojeiro, L.~Padr{\'o}n-Cousillas, S.~Gonzalez-V{\'a}zquez, A.~Carreiro-Alonso, and P.~G{\'o}mez-Area, ``Llms on the fly: Text-to-json for custom api calling,'' \emph{Proceedings of the SEPLN-CEDI}, 2024.

\bibitem{schubotz2016semantification}
M.~Schubotz, A.~Grigorev, M.~Leich, H.~S. Cohl, N.~Meuschke, B.~Gipp, A.~S. Youssef, and V.~Markl, ``Semantification of identifiers in mathematics for better math information retrieval,'' in \emph{Proceedings of the 39th International ACM SIGIR conference on Research and Development in Information Retrieval}, 2016, pp. 135--144.

\bibitem{gallen2024importance}
A.~Gallen, ``The importance of data validation and parsing when working with external data sources,'' 2024.

\bibitem{jiang2021can}
Z.~Jiang, J.~Araki, H.~Ding, and G.~Neubig, ``How can we know when language models know? on the calibration of language models for question answering,'' \emph{Transactions of the Association for Computational Linguistics}, vol.~9, pp. 962--977, 2021.

\bibitem{macedo2024exploring}
M.~Macedo, Y.~Tian, F.~Cogo, and B.~Adams, ``Exploring the impact of the output format on the evaluation of large language models for code translation,'' in \emph{Proceedings of the 2024 IEEE/ACM First International Conference on AI Foundation Models and Software Engineering}, 2024, pp. 57--68.

\bibitem{spreitzer2024mathematical}
C.~Spreitzer, O.~Straser, S.~Zehetmeier, and K.~Maa{\ss}, ``Mathematical modelling abilities of artificial intelligence tools: The case of chatgpt,'' \emph{Education Sciences}, vol.~14, no.~7, p. 698, 2024.

\bibitem{pan2025lemma}
Z.~Pan, Y.~Li, H.~Lin, Q.~Pei, Z.~Tang, W.~Wu, C.~Ming, H.~V. Zhao, C.~He, and L.~Wu, ``Lemma: Learning from errors for mathematical advancement in llms,'' \emph{arXiv preprint arXiv:2503.17439}, 2025.

\bibitem{xu2025large}
F.~Xu, Q.~Lin, J.~Han, T.~Zhao, J.~Liu, and E.~Cambria, ``Are large language models really good logical reasoners? a comprehensive evaluation and beyond,'' \emph{IEEE Transactions on Knowledge and Data Engineering}, 2025.

\bibitem{ashqar2025benchmarking}
H.~I. Ashqar, ``Benchmarking llms for real-world applications: From numerical metrics to contextual and qualitative evaluation,'' \emph{Authorea Preprints}, 2025.

\bibitem{parmar2024logicbench}
M.~Parmar, N.~Patel, N.~Varshney, M.~Nakamura, M.~Luo, S.~Mashetty, A.~Mitra, and C.~Baral, ``Logicbench: Towards systematic evaluation of logical reasoning ability of large language models,'' \emph{arXiv preprint arXiv:2404.15522}, 2024.

\bibitem{jiang2024llms}
Z.~Jiang, H.~Peng, S.~Feng, F.~Li, and D.~Li, ``Llms can find mathematical reasoning mistakes by pedagogical chain-of-thought,'' \emph{arXiv preprint arXiv:2405.06705}, 2024.

\bibitem{luo2024pkrd}
X.~Luo, F.~Ding, Y.~Song, X.~Zhang, and J.~Loo, ``Pkrd-cot: A unified chain-of-thought prompting for multi-modal large language models in autonomous driving,'' \emph{arXiv preprint arXiv:2412.02025}, 2024.

\bibitem{hoffreumon2021multi}
T.~Hoffreumon and O.~Oreshkov, ``The multi-round process matrix,'' \emph{Quantum}, vol.~5, p. 384, 2021.

\bibitem{zhang2022automatic}
Z.~Zhang, A.~Zhang, M.~Li, and A.~Smola, ``Automatic chain of thought prompting in large language models,'' \emph{arXiv preprint arXiv:2210.03493}, 2022.

\bibitem{mirzadeh2024gsm}
I.~Mirzadeh, K.~Alizadeh, H.~Shahrokhi, O.~Tuzel, S.~Bengio, and M.~Farajtabar, ``Gsm-symbolic: Understanding the limitations of mathematical reasoning in large language models,'' \emph{arXiv preprint arXiv:2410.05229}, 2024.

\bibitem{deng2024explicit}
Y.~Deng, Y.~Choi, and S.~Shieber, ``From explicit cot to implicit cot: Learning to internalize cot step by step,'' \emph{arXiv preprint arXiv:2405.14838}, 2024.

\bibitem{liu2023mathematical}
W.~Liu, H.~Hu, J.~Zhou, Y.~Ding, J.~Li, J.~Zeng, M.~He, Q.~Chen, B.~Jiang, A.~Zhou \emph{et~al.}, ``Mathematical language models: A survey,'' \emph{arXiv preprint arXiv:2312.07622}, 2023.

\bibitem{mathews2024test}
N.~S. Mathews and M.~Nagappan, ``Test-driven development for code generation,'' \emph{arXiv preprint arXiv:2402.13521}, 2024.

\bibitem{schoenfeld2010research}
A.~H. Schoenfeld, ``Research methods in (mathematics) education,'' in \emph{Handbook of international research in mathematics education}.\hskip 1em plus 0.5em minus 0.4em\relax Routledge, 2010, pp. 481--533.

\bibitem{gadanidis2017artificial}
G.~Gadanidis, ``Artificial intelligence, computational thinking, and mathematics education,'' \emph{The International Journal of Information and Learning Technology}, vol.~34, no.~2, pp. 133--139, 2017.

\bibitem{liang2024mathematics}
S.~Liang, W.~Zhang, T.~Zhong, and T.~Liu, ``Mathematics and machine creativity: A survey on bridging mathematics with ai,'' \emph{arXiv preprint arXiv:2412.16543}, 2024.

\bibitem{boppana2022machine}
V.~R. Boppana, ``Machine learning and ai learning: Understanding the revolution,'' \emph{Journal of Innovative Technologies}, vol.~5, no.~1, 2022.

\bibitem{shalev2014understanding}
S.~Shalev-Shwartz and S.~Ben-David, \emph{Understanding machine learning: From theory to algorithms}.\hskip 1em plus 0.5em minus 0.4em\relax Cambridge university press, 2014.

\bibitem{alpaydin2020introduction}
E.~Alpaydin, \emph{Introduction to machine learning}.\hskip 1em plus 0.5em minus 0.4em\relax MIT press, 2020.

\bibitem{hastie2009overview}
T.~Hastie, R.~Tibshirani, J.~Friedman, T.~Hastie, R.~Tibshirani, and J.~Friedman, ``Overview of supervised learning,'' \emph{The elements of statistical learning: Data mining, inference, and prediction}, pp. 9--41, 2009.

\bibitem{celebi2016unsupervised}
M.~E. Celebi and K.~Aydin, \emph{Unsupervised learning algorithms}.\hskip 1em plus 0.5em minus 0.4em\relax Springer, 2016, vol.~1.

\bibitem{dike2018unsupervised}
H.~U. Dike, Y.~Zhou, K.~K. Deveerasetty, and Q.~Wu, ``Unsupervised learning based on artificial neural network: A review,'' in \emph{2018 IEEE International Conference on Cyborg and Bionic Systems (CBS)}.\hskip 1em plus 0.5em minus 0.4em\relax IEEE, 2018, pp. 322--327.

\bibitem{li2017deep}
Y.~Li, ``Deep reinforcement learning: An overview,'' \emph{arXiv preprint arXiv:1701.07274}, 2017.

\bibitem{fitz2021neural}
S.~Fitz and P.~Romero, ``Neural networks and deep learning: A paradigm shift in information processing, machine learning, and artificial intelligence,'' \emph{The Palgrave handbook of technological finance}, pp. 589--654, 2021.

\bibitem{lu2019artificial}
Y.~Lu, ``Artificial intelligence: a survey on evolution, models, applications and future trends,'' \emph{Journal of Management Analytics}, vol.~6, no.~1, pp. 1--29, 2019.

\bibitem{team2023gemini}
G.~Team, R.~Anil, S.~Borgeaud, J.-B. Alayrac, J.~Yu, R.~Soricut, J.~Schalkwyk, A.~M. Dai, A.~Hauth, K.~Millican \emph{et~al.}, ``Gemini: a family of highly capable multimodal models,'' \emph{arXiv preprint arXiv:2312.11805}, 2023.

\bibitem{renze2024benefits}
M.~Renze and E.~Guven, ``The benefits of a concise chain of thought on problem-solving in large language models,'' in \emph{2024 2nd International Conference on Foundation and Large Language Models (FLLM)}.\hskip 1em plus 0.5em minus 0.4em\relax IEEE, 2024, pp. 476--483.

\bibitem{yugeswardeenoo2024question}
D.~Yugeswardeenoo, K.~Zhu, and S.~O'Brien, ``Question-analysis prompting improves llm performance in reasoning tasks,'' \emph{arXiv preprint arXiv:2407.03624}, 2024.

\bibitem{chen2025towards}
Q.~Chen, L.~Qin, J.~Liu, D.~Peng, J.~Guan, P.~Wang, M.~Hu, Y.~Zhou, T.~Gao, and W.~Che, ``Towards reasoning era: A survey of long chain-of-thought for reasoning large language models,'' \emph{arXiv preprint arXiv:2503.09567}, 2025.

\bibitem{patil2024review}
R.~Patil and V.~Gudivada, ``A review of current trends, techniques, and challenges in large language models (llms),'' \emph{Applied Sciences}, vol.~14, no.~5, p. 2074, 2024.

\bibitem{badshah2024quantifying}
S.~Badshah and H.~Sajjad, ``Quantifying the capabilities of llms across scale and precision,'' \emph{arXiv preprint arXiv:2405.03146}, 2024.

\bibitem{laskar2024systematic}
M.~T.~R. Laskar, S.~Alqahtani, M.~S. Bari, M.~Rahman, M.~A.~M. Khan, H.~Khan, I.~Jahan, A.~Bhuiyan, C.~W. Tan, M.~R. Parvez \emph{et~al.}, ``A systematic survey and critical review on evaluating large language models: Challenges, limitations, and recommendations,'' in \emph{Proceedings of the 2024 Conference on Empirical Methods in Natural Language Processing}, 2024, pp. 13\,785--13\,816.

\bibitem{liumatheval}
Z.~Liu, T.~Liu, Z.~Chen, M.~Tian, W.~Luo \emph{et~al.}, ``Matheval: A comprehensive benchmark for evaluating large language models on mathematical reasoning capabilities.''

\bibitem{chen2024empirical}
Z.~Chen, Y.~Chen, J.~Han, Z.~Huang, J.~Qi, and Y.~Zhou, ``An empirical study of data ability boundary in llms' math reasoning,'' \emph{arXiv preprint arXiv:2403.00799}, 2024.

\bibitem{lama2024benchmarking}
V.~Lama, C.~Ma, and T.~Ghosal, ``Benchmarking automated theorem proving with large language models,'' in \emph{Proceedings of the 1st Workshop on NLP for Science (NLP4Science)}, 2024, pp. 208--218.

\bibitem{tao2024harnessing}
C.~Tao, X.~Fan, and Y.~Yang, ``Harnessing llms for api interactions: A framework for classification and synthetic data generation,'' \emph{arXiv preprint arXiv:2409.11703}, 2024.

\bibitem{zeng2023evaluating}
F.~Zeng, ``Evaluating the problem solving abilities of chatgpt,'' 2023.

\bibitem{purwono2022understanding}
P.~Purwono, A.~Ma'arif, W.~Rahmaniar, H.~I.~K. Fathurrahman, A.~Z.~K. Frisky, and Q.~M. ul~Haq, ``Understanding of convolutional neural network (cnn): A review,'' \emph{International Journal of Robotics and Control Systems}, vol.~2, no.~4, pp. 739--748, 2022.

\bibitem{schmidt2019recurrent}
R.~M. Schmidt, ``Recurrent neural networks (rnns): A gentle introduction and overview,'' \emph{arXiv preprint arXiv:1912.05911}, 2019.

\end{thebibliography}

\section*{Appendix}

\subsection{Examples from mathematical problems}

\begin{figure*}[htbp]
    \centering
    \includegraphics[width=\linewidth]{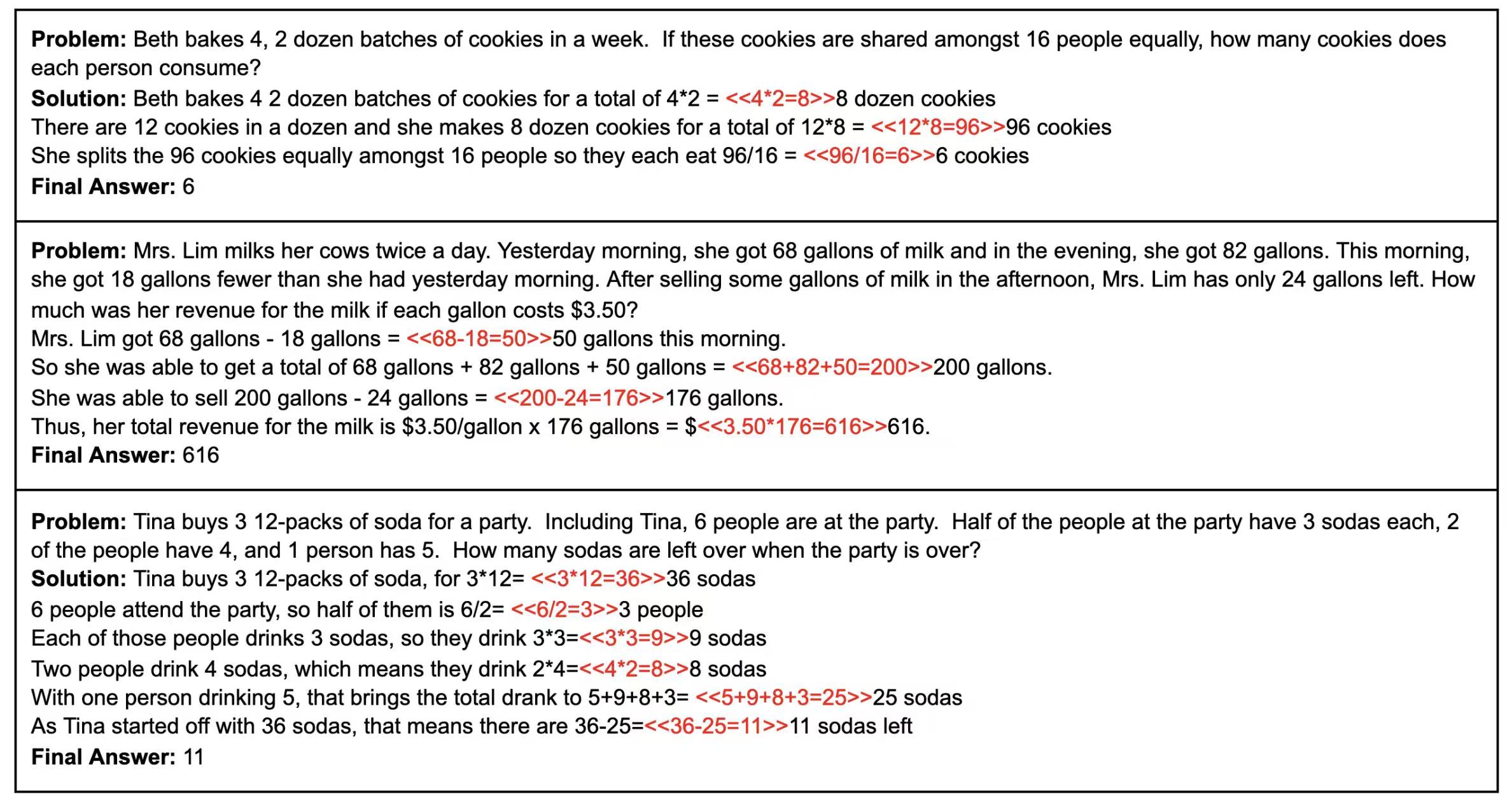}
    \caption{Selected questions and LLM answers from the GSM8K dataset}
    \label{fig:sample_question_answer_gsm8k}
\end{figure*}

\begin{figure*}[htbp]
  \centering
  \begin{subfigure}[b]{1\linewidth}
    \includegraphics[width=0.61\linewidth]{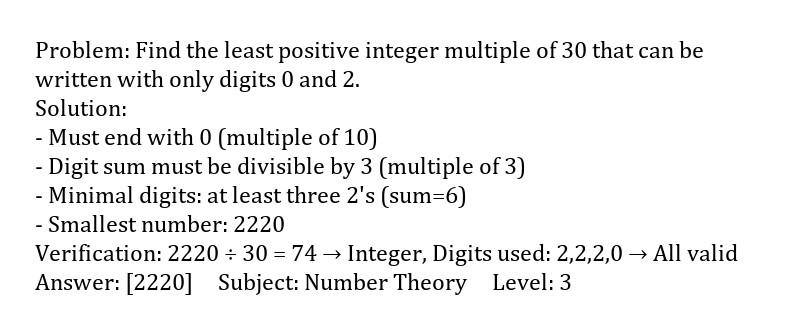}
    \caption{MATH500 example 1}
  \end{subfigure}
  \hfill
  \begin{subfigure}[b]{1\linewidth}
    \includegraphics[width=0.71\linewidth]{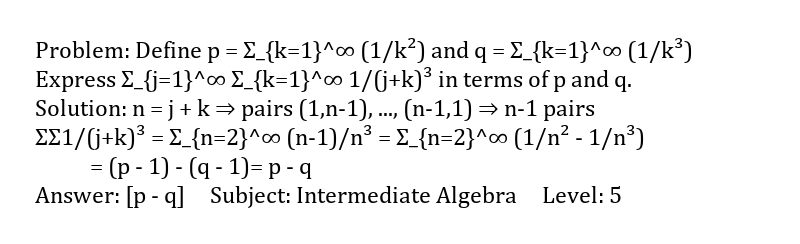}
    \caption{MATH500 example 2}
  \end{subfigure}
  \caption{Selected examples from the MATH500 dataset}
  \label{fig:math500_examples}
\end{figure*}

'


\end{document}